\documentclass{article}

     \PassOptionsToPackage{numbers, compress}{natbib}

\usepackage{multirow}
\usepackage{times}
\usepackage{latexsym}
\usepackage{adjustbox}
\usepackage{wrapfig}
\usepackage{booktabs}
\usepackage{makecell}
\usepackage{longtable}
\usepackage{CJKutf8}
\usepackage{pdfpages}
\usepackage[T1]{fontenc}

\usepackage[utf8]{inputenc}

\usepackage{microtype}

\usepackage[preprint]{neurips_2023}

\usepackage[utf8]{inputenc} 
\usepackage[T1]{fontenc}    
\usepackage{hyperref}       
\usepackage{url}            
\usepackage{booktabs}       
\usepackage{amsfonts}       
\usepackage{nicefrac}       
\usepackage{microtype}      
\usepackage{xcolor}         
\usepackage{amsmath,amssymb}

\title{Zero-shot Conversational Summarization Evaluations with small Large Language Models}

\author{%
  Ramesh Manuvinakurike \\
  Intel labs, USA \\  \texttt{} \\
  \And
  Saurav Sahay \\
  Intel labs, USA\\  
  \texttt{} \\
  \AND
  Sangeeta Manepalli \\
  Intel, USA\\  
  \texttt{} \\
  \And
  Lama Nachman \\
  Intel labs, USA\\  
  \texttt{} \\
}

\begin{document}

\maketitle

\begin{abstract}
  Large Language Models (LLMs) exhibit powerful summarization abilities. However, their capabilities on conversational summarization remains under explored. In this work we evaluate LLMs ($\sim$10 billion parameters) on conversational summarization and showcase their performance on various prompts. We show that the summaries generated by models depend on the instructions and the performance of LLMs vary with different instructions sometimes resulting steep drop in ROUGE scores if prompts are not selected carefully. We also evaluate the models with human evaluations and discuss the limitations of the models on conversational summarization. 
\end{abstract}

\section{Introduction}

Large Language Models (LLMs) have been shown to exhibit impressive multitasking abilities including summarization \cite{kojima2022large}. 
These LLMs show great promise for automatic summarization and achieve impressive zero-shot performance \cite{zhang2023benchmarking,wang2023element}. 
LLMs are `prompted' to solve a task such as summarization by issuing an instruction (e.g., Generate a summary for this dialogue, Summarize: etc.) and an input (conversation, news article2, queries etc.). These instructions are rather simple natural language phrases appended as a prefix or postfix to the input to assist the LLMs in carrying out the task. 
Generating such instructions and prompting can be challenging for novices \cite{zamfirescu2023johnny}. 
The brittleness of the LLMs in processing the prompts consistently is one such reason that causes a challenge \cite{zhao2021calibrate,webson2022prompt,lu2022fantastically,kaddour2023challenges}. 
A minor variation in the prompt syntax can result in vastly different results. Prompt engineering has thus evolved to design approaches for the LLMs to produce the desired results consistently. Several resources exist for prompt engineering. Promptsource is one such popular repository consisting of 2000+ prompts for natural language prompts for various natural language tasks including summarization. \cite{bach2022promptsource}. 
In spite of these advances, the brittleness exist and more so in smaller open-sourced LLMs. 
One possible solution to tackle the brittleness could be to further finetune the models using the instructions for a task coupled with data points from a dataset at the developer's disposal. This however could result in loss of generalization abilities of the LLMs even within a task (for instance, finetuning on conversational summarization from a technical design meeting dataset could result in drop in summarization capabilities on chit-chat data). In this work, we finetune the LLMs on a technical council meetings dataset and evaluate the performance on a chit-chat dataset 
and discuss the results. 

This paper arises from a need for evaluating open-sourced LLMs \cite{kaddour2023challenges} (within $\sim$ 11 billion parameters) for conversation summarization and their robustness to different prompts (The rationale for selecting models within $\sim$11b parameters is discussed in detail in Sec~\ref{subsec:param_count}). 
These specific sized LLMs provide the capability run at the edge on PCs or mobile systems preserving the privacy of the user data while benefiting the user with productivity gains.
These models need to be benchmarked against the existing conversational summarization datasets by measuring their abilities on different prompts. We motivate this work with the following research questions: 
\\
1) How important is prompt selection for conversational summarization task? How do LLMs less than $\sim$11 billion parameters fare on different prompts varying slightly in syntactic structure? How do these models fare in human evaluation?
\\
2) Can finetuning the LLMs for conversational summarization a task but on a different domain improve zero shot summarization capabilities? 
\\
We also discuss some limitations of the models specially focusing on hallucinations and role attributions in the summaries. 

\section{Related work}

Recent work on benchmarking the LLM abilities for news articles summarization \cite{zhang2023benchmarking} is the most related our work. The authors highlight there exists a fundamental limitation with the existing news articles summarization datasets and found that the target/gold-truth summaries were sub par for comparative evaluations. To overcome the problem the authors hired freelance writers to generate target summariesand pitch the summaries generated by the models are pitched against those written by expert freelancers. The authors found that the summaries generated by the models were on-par with the summaries generated by the experts \cite{zhang2023benchmarking} which in turn is better than the target summaries. It is important to note that the LLMs used in this work range in parameter count from 350M to 175 billion \cite{brown2020language,du2021glm,ouyang2022training,zhang2022opt,bai2022training}. In this work, we're focused on benchmarking the LLMs on the task of conversational summarization.

It has been well established that training the models on instructions improves the generalization abilities in the LLMs (see for e.g., \cite{wei2021finetuned,mishra2022cross,sanh2021multitask,brown2020language,ouyang2022training,hase2022can,zhong2021adapting}. Several resources such as Promptsource\cite{bach2022promptsource}, Natural Instructions \cite{mishra2022cross}, Flan \cite{wei2021finetuned} and, Super Natural Instructions \cite{supernaturalinstructions} have been developed in recent times to provide researchers and engineers alike with resources to study, find and leverage instructions for a task of interest. In this work, we choose the prompts from Promptsource \footnote{
we study Promptsource in this work due it's high activity and star ratings (compared to similar resources).
}.

\section{Data}

Several promising datasets exist for the task of conversational summarization \cite{elitr-minuting-corpus:2022,zhong2021qmsum,carletta2005ami,janin2003icsi,chenetal2021dialogsum,hu2023meetingbank,gliwa2019samsum}. 
In this work we utilize the MeetingBank dataset \cite{hu2023meetingbank} for instruction tuning purposes.
The MeetingBank dataset is relatively new and serves as a great resource for evaluating the finetuning impacts on the models on a specialized domain.
The dataset contains transcriptions from various United States cities or municipality meetings. We then evaluate the models in a zero-shot manner on Samsum corpus \cite{gliwa2019samsum} and Dialogsum datasets \cite{chen-etal-2021-dialogsum}. We utilize the validation splits from Huggingface datasets repository for our study purposes. 

For the purpose of finetuning the model we split the Meetingbank dataset into 95\% training and 5\% validation split and utilize the training set for instruction tuning. 
For the purposes of instruction tuning we utilize Promptsource\cite{bach2022promptsource} which is a crowdsourced valuable resource consisting of thousands of prompts contributed for numerous NLP tasks and datasets. 
We utilize the prompts for conversational summarization section (Table~\ref{tab:prompts}). 
We further instruct-tune the models using a pipeline similar to T0 \cite{sanh2021multitask}. 
For training \& inferencing each transcription input is converted to a prompt for LMMs by `apply'ing the template with the conversations using Langchain \cite{chaselangchain}.
 
\section{Models}

Conversational summarization has increasingly gained interest in a post-covid world with the nature of workplace becoming more hybrid \cite{hybrid-work-mckinsey}. This has amplified the need for a conversational/meeting summarization systems to run on machines with limited compute capabilities with democratization aspirations. 

\subsection{Choice for parameter counts}
\label{subsec:param_count}
Our choice of models with $\sim$11 billion parameters is mainly motivated by the constraint imposed by the cost and, hardware limitations and a motivation to study less-carbon emitting approach of model development. These model sizes are also conducive to running on PCs and mobile systems, however there is an added impact of bigger models and more data needed for inferencing which is observed in the results noted below in the Hardware Inference section using an LLM as an example workload. 
\textbf{Hardware Training:} Fine-tuning a 7 billion parameter model takes approximately 66 GB of memory (RAM) at 32 bit precision and a batch size of 1 (sequence length of 2048) (see memory computation in \cite{korthikanti2023reducing}) which is slightly less than a high end GPU available in the market. While several optimization approaches exist which significantly reduce the compute requirements (e.g., Quantization \cite{frantar2022gptq}), these are accompanied by often deteriorated performance or comparable in the best case. Running an inference locally on client devices (without use of cloud APIs/services) even with quantization of models is memory intensive \footnote{https://github.com/ggerganov/llama.cpp}. While bigger is often better in LLMs, it is important to study the trade-offs in models whose parameter counts are still tractable and can be deployed for inferencing (and even train) using reasonable compute budgets. 
\textbf{Hardware Inference:} Running inference instances is also an important consideration. For instance, we consider a GPT variant\footnote{https://github.com/imartinez/privateGPT} on a state-of-the-art PC (Intel Raptor Lake 28W Core i7 system (1380P) configured with 32GB of memory and 512 GB of storage). For running an inferencing pipeline that includes document ingestion, encoding,
followed by autoregressive decoding 
with a ~7B parameter model (
\cite{gpt-j}) generated a bandwidth of ~10 GB/s. First inference with model loading was significantly slower $\sim$0.5 FPS. 
About 1/3rd of memory was consumed by text encoding and LLM operation and reduced the headroom for other applications. 
Optimizations to use compressed models with reducing memory interactions is the key to use these models effectively on PCs and mobile systems. Models with any greater parameter count would pose additional inferencing challenges. \textbf{Carbon emissions:} Training a 70B emits 833\% more carbon and 13B parameter model emits 100\% more carbon than the 7B parameter model. While, the developing organization offset the carbon, it might not be feasible for all\cite{touvron2023llama2}. 
With these experiments \& observations we arrived at our model evaluations to those mentioned below. In the future we plan to extend the evaluations to models of greater parameter count.  

\subsection{Models selected}
\textbf{Llama } is a foundational language model developed with publicly available datasets by Gen AI, Meta. Llama 2 \cite{touvron2023llama2} is a more recently released open source class of foundational large language models pretrained \& finetuned with more data and higher context lengths than it's predecessor Llama 1\cite{touvron2023llama}. Llama 2 models achieve impressive performance on numerous public benchmarks. We choose Llama 2 as one of the models for our benchmarking purposes. The family also includes Llama-2-chat model which is developed using supervised finetuning and then iteratively refined using Reinforcement Learning with Human Feedback (RLHF) \cite{ouyang2022training}. 
\textbf{Falcon} \cite{falconllm} is a family of generative LLM by Technology Innovation Institute which achieves impressive performances on many of the LLM benchmarks. The models are trained predominantly on RefinedWeb \cite{penedo2023refinedweb} along with other data from the web. The instruct version of the models are further finetuned on instructions and conversational data. 
\textbf{Alpaca} \cite{alpaca} is a instruction-tuned model based on Llama \cite{touvron2023llama} which is instruction tuned using the techniques from self-instruct \cite{wang2022self} paradigm. The alpaca model was found to exhibit behavior similar to text-davinci-003 model \cite{Openaichatgpt}. 
\textbf{Flan-T5} are a class of language models \cite{chung2022scaling} where encoder-decoder transformer-based model \cite{raffel2020exploring} are finetuned on 1836 tasks using instruction-finetuning and chain-of-thought finetuning paradigms.  The models exhibit state-of-the-art abilities in various benchmark tasks. 
We also benchmark the results using \textbf{GPT3.5 turbo}  \cite{Openaigpt35turbo} which is a an improved GPT-3 model with instruction following ability on Natural language and code. While, the parameter count of the model is in hundreds of billion, the purpose of including the model is for comparison. 

\section{Experiments}

Our goal is compare the model performance on conversational summarization (Samsum \& Dialogsum) and then evaluate the model performance on different prompts. There are 6 prompts in Promptsource for conversational summarization (Table ~\ref{tab:prompts}) that we then apply using Langchain on the each conversation as input to the models. We then extract the output from these models as summaries. 
We additionally finetune existing instruction-tuned models- (Flan \& Lama2-7b-chat) using task specific instructions and the data from Meetingbank dataset (Models marked as **). 
While instruction tuning has been shown to improve the overall performance in the model generalization abilities \cite{mishra2022cross}, we intend to explore if the task performance (conversational summarization) utilizing the data from a related but different dataset can further provide a boost. 

We report Rouge scores (R1, R2, RL) \cite{lin2004rouge} and BertScore \cite{zhang2019bertscore} on the summaries generated by the models comparing against the available ground truth summaries. We additionally report Length Ratio (LR), Novel uni/bi (N1, N2) gram metrics which can be evaluated automatically on the Dialogsum and the Samsum datasets which the models haven't seen during our training process \footnote{T0pp model however have been exposed to Samsum dataset off-the-shelf. 
We cannot however guarantee that the models have not been exposed to these datasets due to limited information available. }. We compute Length Ratio = length (tokens in summaries generated)/length (tokens in the conversations), N-gram novelty = length (N-gram tokens in summaries generated $\cap$ tokens in conversation)/ length(tokens in the conversations).

Target summaries don't always serve as a reliable ground truth for  evaluation \cite{zhang2023benchmarking}. 
Summarization task is subjective and also information can be presented in several different forms (sentences \& format). However, the summaries in the dataset themselves can be relied upon as one of the many possible summaries for a given conversation. We thus experiment with humans to compare the model generated summary with the target and report the findings here. 
We developed a web-interface (Figure~\ref{fig:ratings_ui}) for collecting human preferences between target and model generated summary. The conversation is displayed to the users along with two summaries, one model generated and the other reference/target summaries. The summary orders are jumbled and the raters are not aware of the source of summaries. The raters were volunteers and employees of anonymous and had education levels of graduate degree and were aware of the task. Each user was requested to rate as many summaries as they could. We obtain comparative ratings for Overall, Novelty, Informativeness and Conciseness of the summaries. 
To further understand if the summaries contains hallucinations, correct role attribution and if the summaries had misrepresented gender attributes we developed a similar interface, but this time requesting the human evaluators to rate only the generated summaries if they consist hallucinations, if the roles are properly attributed in the summaries and if the genders are properly attributed in the summary (if applicable). 

\section{Results}
\begin{wrapfigure}{r}{0.66\textwidth}
  \begin{center}
    \includegraphics[width=0.63\textwidth]{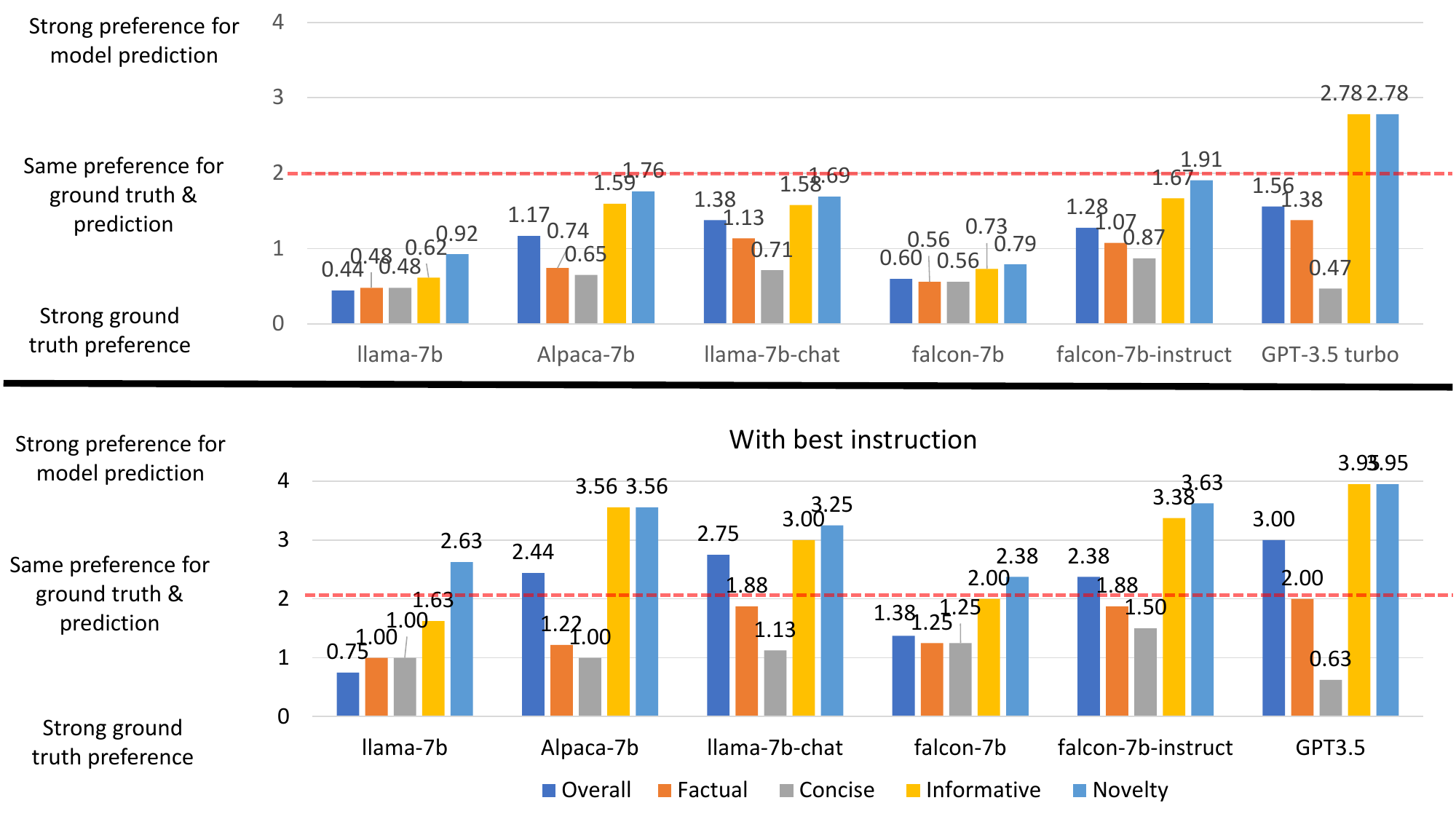}
  \end{center}
  \label{fig:hum_ratings_llm}
  \caption{ Shows the evaluation of the LLMs on various metrics for conversational summarization with any instruction from Promptsource vs. oracle selected best instruction.  }
\end{wrapfigure}
\textbf{Do Instruction tuned models perform generate better summaries?} We can observe from Table~\ref{tab:results} that when comparing the instruction tuned models (Alpaca-7b vs Llama-7b, Falcon-7b-instruct vs Falcon-7b) with just pre-trained model, the instruction tuned models generate summaries with better Rouge-scores, BertScores, more concise and generate fewer novel 1-gram and 2-gram (due to fewer hallucinations). We also find that the summaries generated by the instruction tuned models are significantly better w.r.t Rouge scores and BERT scores (p < 0.001, Wilcoxon rank-sum test) than the summaries generated by the non-instruction tuned models.
To arrive at this conclusion we compared sample-wise Rouge \& BERT scores in separate experiments between the non-instruction tuned models and their instruction tuned counterparts. This implies that instruction tuning helps improve the conversational summarization performance which is consistent with the literature (on a different task) \cite{mishra2022cross}.
Humans also rated the summaries from instruction tuned model higher than pre-trained models (Figure 1
). 

\textbf{Does further task-specific instruction tuning help models perform better?} We find that further 
 \begin{wrapfigure}{r}{0.6\textwidth}
  \begin{center}
    \includegraphics[width=0.58\textwidth]{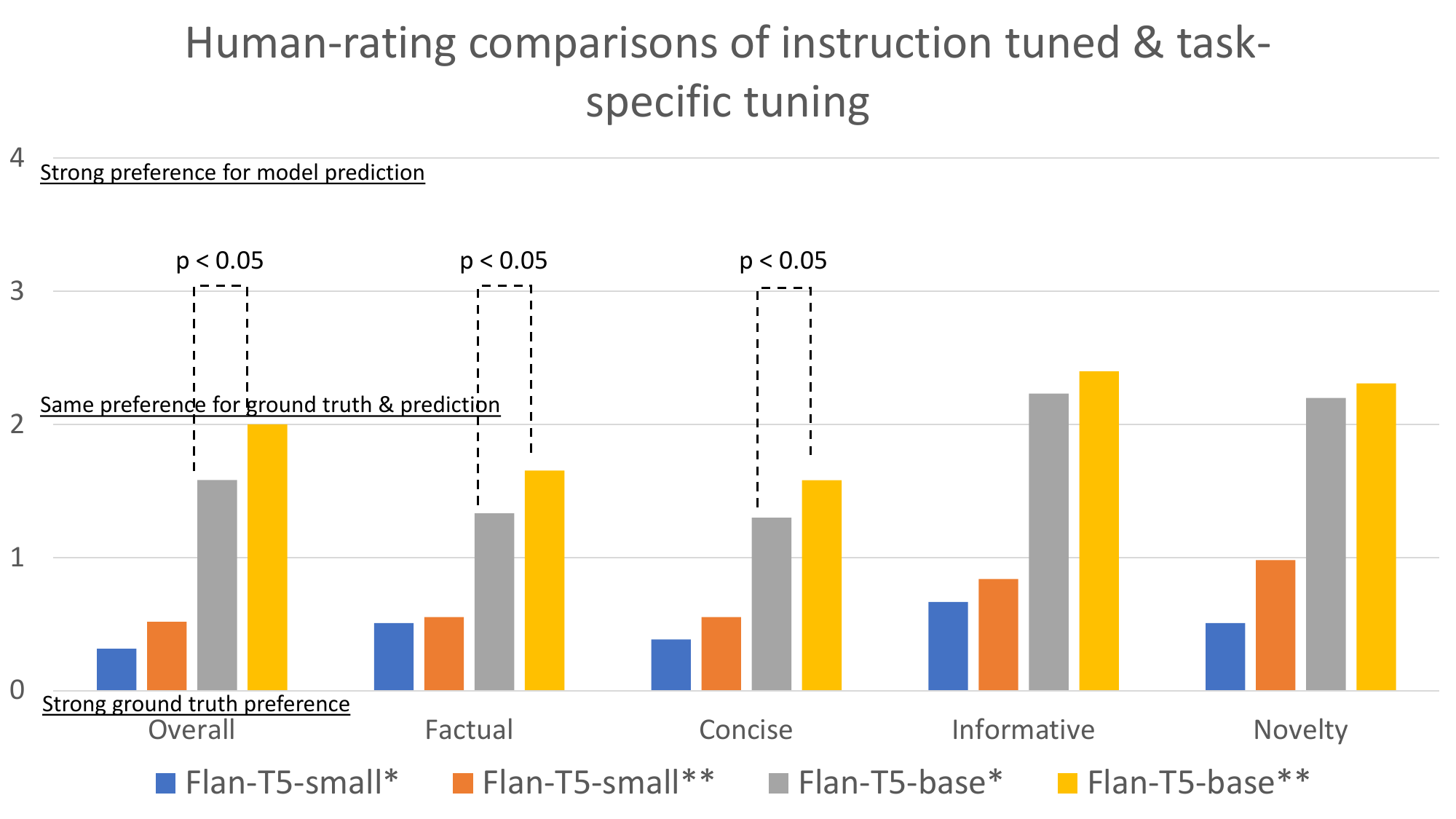}
  \end{center}
  \label{fig:finetuning_hum_ratings}
  \caption{Shows the human evaluations on various metrics comparing instruction tuned model * and task-specific finetuned model ** demonstrating improvements.}
\end{wrapfigure}
 tuning the models for a task (summarization) even on a different dataset (Meetingbank) slightly improves the performance of the models. In Table~\ref{tab:results} we identify task-specific instruction tuned model with `**' vs available off-the-shelf instruction tuned checkpoints `*' (Flan-T5-\{MODEL\_SIZE\}* vs  Flan-T5-\{MODEL\_SIZE\}** , Llama-7b-chat* vs Llama-7b-chat**). We do not perform any experiments to verify if the model's performance on a more general abilities since our goal is to leverage LLMs for summarization task alone. This slight change in the summary quality was also resulted in increase in preferences over target in human-evaluation overall and even on their factual and conciseness (Flan-T5-base* vs. Flan-t5-base**, Flan-T5-small* vs. Flan-T5-small** in Figure 2
 ). This is important, since we're not pitching the models one against the other, but rather against the target summaries. 

\textbf{Do summary qualities differ across instructions and models?} We also find that the model's summarizing abilities are highly dependent on the instruction used and the model itself. For instance, `Given the above dialogue write a summary' as an instruction gives 8\% better Rouge-L scores than `Generate a summary for this dialogue' in Falcon-7b instruct tuned model compared to `Llama-7b' which gives the best Rouge scores from the prompt `Given the above dialogue write a summary'. We also tested this variable performance using ANOVA test (single factor, p < 0.0001 (F=34.56, df=6, MS=0.029) and find that the difference in Rouge scores are significantly different across prompts. Figure 3
shows R-L box-plots for each prompt on SAMSUM corpus. We further find that the variability in instruction tuned models is slightly lower compared to non-instruction tuned models. 
This result highlights the need for future work to reduce the gap between the model performance across prompts which could make it easier for non-experts to query the LLMs. 
We also find that this variability reduces with increase in model parameters. 

\begin{wrapfigure}{r}{0.62\textwidth}
  \begin{center}
    \includegraphics[width=0.58\textwidth]{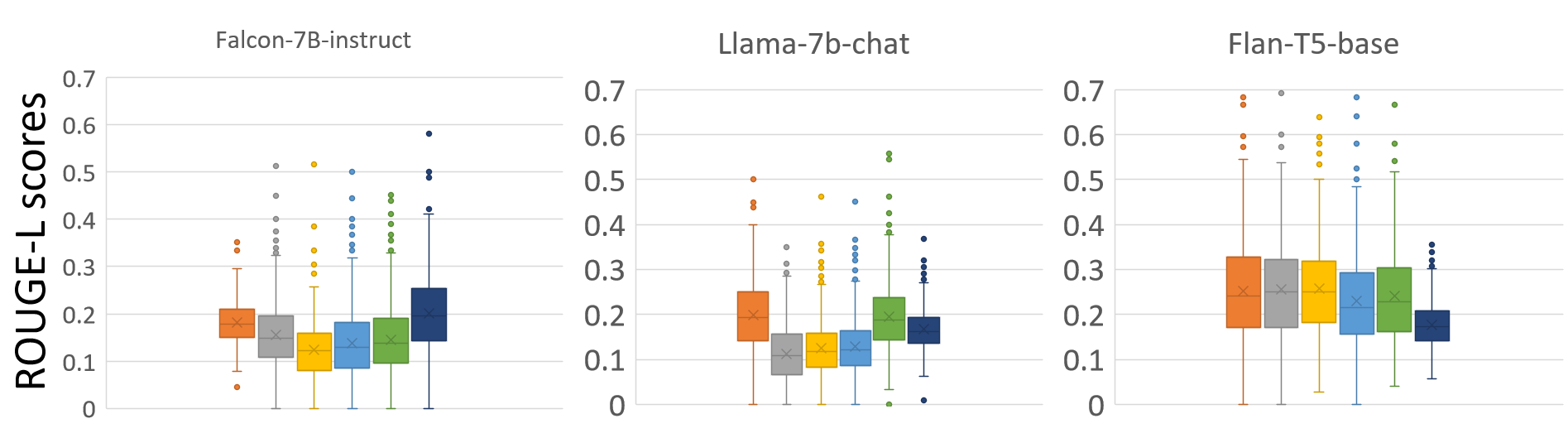}
  \end{center}
  \label{fig:prompt_var}
  \caption{Box plot of Rouge scores of each sample from Samsum dataset showing variability for each prompt (Table~\ref{tab:prompts}). Each color represents different prompt.  }
\end{wrapfigure}

\textbf{If we were to use the best prompt, how do LLMs compare with target summaries?} We selected the prompt yielding the highest Rouge-L scores (oracle selection) and then compare the human ratings for the best performing prompt for each of the model. Upon selecting the best prompt we find that all the LLM generated summaries are rated more novel than the target summaries (Figure 1
under `With best instruction'). All the instruction tuned models (Alpaca, Llama-chat, Falcon-instruct) were rated more informative and were preferred overall compared to the target summaries. However, the target summaries are rated most factually consistent and concise compared to the generated summaries. Lower rating for the conciseness were more obvious when we compare the LR of the generated summaries vs the target summaries (Dailogsum target summaries LR = 14.85\%, Samsum LR = 17.94\%) compared to the summaries length of the model generated summaries. Lower factual ratings can be explained by presence of hallucinations and other errors. Table~\ref{tab:results_finetuning} shows best-possible Rouge and BERT scores on the oracle selected best prompts used once per dataset. 

\begin{table*}[]
    \centering
    \begin{adjustbox}{width=\textwidth}
    \begin{tabular}{| l   
                    | c | c | c | c | c | c | c
                    | c | c | c | c | c | c | c | }
    \hline
            \multirow{1}{*}{Model} 
                & \multicolumn{7}{c|}{Samsum} 
                & \multicolumn{7}{c|}{Dialogsum}
                \\ \cline{2-15}
                & R-1 & R-2 & R-L & B-Score & LR & N1 & N2
                & R-1 & R-2 & R-L & B-Score & LR & N1 & N2
                \\ \hline
                
        Llama-7b &  
                  0.14 & 0.02 & 0.10 & 0.701 & 56.58 & 47.06 & 85.70 &
                  0.16 & 0.02 & 0.11 &  0.736 & 40.24 & 31.22 & 61.87
                  \\ \hline
        Falcon-7b &  
                   0.14 & 0.02 & 0.10 & 0.703 & 58.80 & 48.74 & 89.68 &
                   0.16 & 0.02 & 0.11 &  0.737 & 36.32 & 26.61 & 54.37
                   \\ \hline
        \hline        
        Alpaca* & 
                0.18 & 0.04 & 0.13 & 0.721 & 35.19 & 27.82 & 52.80 &
                0.17 & 0.03 & 0.13 &  0.740 & 21.78 & 15.06 & 32.29
                \\ \hline
        Falcon-7b-inst* &  
                    0.21 & 0.05 & 0.15 & 0.739 & 33.96 & 27.82 & 52.63 &
                    0.22 & 0.05 & 0.15 & 0.761 & 24.99 & 18.10 & 38.80
                    \\ \hline  
           \hline
               
        Flan-T5-small* &   
                  0.42 & 0.19 & 0.33 & 0.844 & 24.38 & 8.71 & 27.25 &
                  0.22 & 0.07 & 0.17 & 0.742 & 17.11 & 5.54 & \textbf{15.62}
                  
                  \\ \hline
        Flan-T5-small** &   
                   0.43 & 0.20 & 0.33 & 0.845 & 24.45 & \textbf{8.70} & 27.33 &
                   0.23 & 0.07 & 0.18 & 0.741 & 17.16 & 5.63 & 15.86
                   \\ \hline \hline
        Flan-T5-base* &                  
                 0.46 & 0.23 & 0.36 & 0.856 & 24.29 & 9.14 & 28.01 &
                 0.31 & 0.11 & 0.24 & 0.785 & 18.82 & \textbf{4.80} & 15.93
                 \\ \hline
        Flan-T5-base** &                  
                0.47 & 0.23 & 0.37 & 0.857 & 24.23 & 9.09 & 27.92 &
                0.32 & 0.12 & 0.25 & 0.787 & 18.58 & 4.82 & 15.97
                 \\ \hline \hline
                 
        Flan-T5-large* &  
                  0.44 & 0.19 & 0.35 & 0.849 & 18.38 & 10.24 & 25.25 &
                  0.29 & 0.09 & 0.23 & 0.793 & 13.23 & 7.59 & 18.75
                  \\ \hline
         Flan-T5-large** &  
                 0.46 & 0.20 & 0.36 & 0.851 & 18.38 & 10.24 & 24.22 &
                 0.30 & 0.10 & 0.24 & 0.793 & 13.24 & 7.52 & 18.72 
                  \\ \hline \hline
                      
        Flan-T5-xl* &  
               0.48 & 0.26 & 0.39 & 0.862 & 24.37 & 9.87 & 28.21 &
               0.37 & 0.16 & 0.29 & 0.824 & 18.10 & 5.71 & 18.13
               \\ \hline 
        Flan-T5-xl** &  
               0.49 & 0.26 & 0.40 & 0.863 & 24.41 & 9.92 & 28.27 &
               0.38 & 0.17 & 0.31 & 0.827 & 18.17 & 5.71 & 18.13 
               \\ \hline \hline

        Llama-7b-chat* &  
                       0.18 & 0.04 & 0.13 & 0.721 & 32.41 & 25.65 & 48.58 &
                       0.21 & 0.05 & 0.15 &  0.765 & 31.46 & 22.54 & 47.39
                       \\ \hline 
        Llama-7b-chat**$^\pm$ &  
               0.18 & 0.05 & 0.14 & 0.721 & 31.66 & 25.03 & 47.45 &
               0.21 & 0.06 & 0.15 & 0.760 & 30.54 & 22.17 & 46.17 
               \\ \hline  \hline       
        GPT-3.5 turbo &  
               0.14 & 0.06 & 0.10 & 0.746 & 189.27 & 80.86 & 195.89 &
               0.14 & 0.05 & 0.11 & 0.741 & 140.31 & 62.33 & 150.97                
               \\ \hline  \hline
    \end{tabular}
    \end{adjustbox}
    \caption{Evaluation of models on the summarization task. * indicates the models are instruction tuned, ** means the models were task instructed tuned for summarization task. Llama-7b-chat was finetuned using LORA\cite{hu2021lora}.}
    \label{tab:results}
\end{table*}


\section{Discussions \& Future work}

\textbf{Do low human ratings for factual ratings correlated to hallucinations and role attributions?}  Hallucinations a challenge to be addressed in LLMs. We collected human ratings to further analyse the cause for lower rating for the factual ratings in LLMs. We requested human raters to mark (Yes/Now) if the summary contained irrelevant phrases not related to the input. We sampled 286 summaries across models randomly from the dataset for this small scale study. We also requested human raters to answer yes/no question if the the summaries had incorrect role attributions and if there was assign/misrepresented gender pronouns. We find that the presence of hallucinations is strongly negatively correlated to the ratings of factual and overall ratings (PCC, $\rho$= -0.77, $\rho$= -0.71 resp.). The ratings are also weakly/moderately negatively correlated to incorrect speaker attributions ($\rho$= -0.31). The raters were more forgiving to incorrect gender attributions (he instead of she) towards overall summary rating ($\rho$= -0.18). This means that hallucinations and incorrect role attributions could've played a more important role in lower factual and overall ratings. Since, our sample size was low we cannot conclude with conviction. This remains an interesting area of exploration. 

\textbf{Are the models with higher overlap with target summaries or input conversations rated more informative?} Rouge (p) between the input and summary generated measures what fraction of the conversation was covered in the summary. We measured the Pearson's correlation co-efficient (PCC) between these Rouge (p) scores (Table~\ref{tab:copying_p}) and the human preference ratings to measure if any correlation exists. We find moderate correlation (PCC, $\rho$ = 0.46) for the informativeness between the models Rouge scores and the preference ratings. This implies that the raters tended to mark models with higher input coverage as more informative. Additionally, we also found that high rouge (p) scores correlated moderately with Overall and Factual ratings. 
This was also true for informativeness. This is however not all that surprising but a reinforcing indicator that summaries need to be concise while increasing the overall coverage. 

Higher novelty ratings (`Select the summary which you think is a more novel way to present the conversation') was an  interesting observation. The LLMs consistently performed better on novelty preferences compared to the target. We believe that novel ways of presenting the summaries (e.g., as question answering exchanges, shorter rephrased conversation versions, alternative narrations) along with novel phraseology could be contributing towards higher novelty ratings. We additionally note that the human raters being experts (have good knowledge of technology and recent advancements) rather than novices could've contributed to this. We intend to explore this in our future work.


Conversational summarization is an important application area of the LLMs and remains an interesting area of research. In this work we evaluated various LLMs on conversational summarization datasets and showed that the instruction tuning yields better results. Given the parameter sizes of the LLM models evaluated, this outcome has interesting applications to deployment of <11B parameter LLMs on PC and mobile systems. We also showed that further task-specific finetuning could yield more promising results. We further analysed the summaries via human and automatic evaluations and discussed the limitations including hallucinations and incorrect role attributions. 


\bibliographystyle{plainnat}
\bibliography{anthology,custom}

\appendix

\section{Appendix}

\subsection{More Related works}
Evaluating LLMs remain a challenge. Promising benchmarks such as HELM \cite{liang2022holistic} have been developed to evaluate the LLM abilities. While news summarization features in the benchmark, conversational summarization is missed out. In this we also argue for inclusion of conversational summarization in the benchmark. Automatic evaluation of summarization remains a challenge \cite{fabbri2021summeval}. 
ROUGE \cite{lin2004rouge} and BERTScore \cite{zhang2019bertscore} still remain the most commonly reported reports for benchmarking purposes. Other metrics compute the quality of the summaries based on  Factual consistency \cite{kryscinski2019evaluating}. Given, the subjective nature of the summary and the difficulty of developing a single or even multiple target summaries, reference free metrics have been developed in recent times to compute the quality of the summaries \cite{vasilyev2020fill}. 
Human evaluations still remain central to computing the quality of the summaries \cite{khashabi2021genie}. Other metrics rely on question-answering approaches to measure the quality of the summaries \cite{scialom2019answers}. 

\subsection{Other LLMs}
\textbf{MPT-7b} MPT \cite{MosaicML2023Introducing} is a family of generative LLM from MosaicML NLP team trained on 1T tokens of text and code. The performance of the model is comparable to Llama \cite{touvron2023llama}. The instruct version of the model is further trained on for instruction following from a dataset \cite{mosaicml2023dolly_hhrlhf} derived from \cite{DatabricksBlog2023DollyV2} and \cite{bai2022training}. 
\\
\textbf{StableLM-7b} is a decoder only LLM based on GPT-NeoX \cite{gpt-neox-library} from Stability AI. The model is trained partly on RefinedWeb extract \cite{penedo2023refinedweb}, RedPajama data \cite{together2023redpajama} and, The Pile dataset \cite{gao2020pile}. 
\\
\textbf{Dolly} Dolly is a class of LLMs based on Pythia \cite{biderman2023pythia} trained on Databricks 15k dolly dataset \cite{DatabricksBlog2023DollyV2} and The Pile \cite{gao2020pile} dataset. From the authors claim Dolly is not a state-of-the-art model, however has good instruction following capabilities. 
\\
\textbf{T0pp} T0 is a class of encoder-decoder models based on T5 \cite{raffel2020exploring} trained in a multi-task prompted setting consisting of various NLP tasks with the goals of improving zero-shot generalizational abilities \cite{sanh2021multitask}. T0pp exhibits strong generalizational abilities on various tasks including summarization. 
\\

\begin{table*}
    \centering
    \begin{adjustbox}{width=\textwidth}
    \begin{tabular}{| l   
                    | c | c | c | c | c | c | c
                    | c | c | c | c | c | c | c | }
    \hline
            \multirow{1}{*}{Model} 
                & \multicolumn{7}{c|}{Samsum} 
                & \multicolumn{7}{c|}{Dialogsum}
                \\ \cline{2-15}
                & R-1 & R-2 & R-L & B-Score & LR & N1 & N2
                & R-1 & R-2 & R-L & B-Score & LR & N1 & N2
                \\ \hline
                
        Llama-7b &  
                  0.14 & 0.02 & 0.10 & 0.701 & 56.58 & 47.06 & 85.70 &
                  0.16 & 0.02 & 0.11 &  0.736 & 40.24 & 31.22 & 61.87
                  \\ \hline
        Falcon-7b &  
                   0.14 & 0.02 & 0.10 & 0.703 & 58.80 & 48.74 & 89.68 &
                   0.16 & 0.02 & 0.11 &  0.737 & 36.32 & 26.61 & 54.37
                   \\ \hline
        MPT-7b &  
                0.14 & 0.02 & 0.10 & 0.693 & 58.30 & 45.26 & 83.00 &
                0.16 & 0.02 & 0.11 &  0.734 & 44.80 & 34.93 & 68.13
                \\ \hline
        Stablelm-7b &  
                         0.14 & 0.01 & 0.10 & 0.691 & 56.45 & 45.86 & 83.52 &
                         0.14 & 0.02 & 0.10 & 0.732 & 42.77 & 33.50 & 66.29
                         \\ \hline \hline
        MPT-7b-inst* &  
                         0.15 & 0.03 & 0.11 & 0.703 & 60.57 & 49.44 & 88.52 &
                         0.16 & 0.02 & 0.11 &  0.729 & 44.57 & 40.49 & 74.74
                         \\ \hline
        MPT-7b-chat* & 
                     0.17 & 0.04 & 0.12 & 0.713 & 63.64 & 49.92 & 92.34 &
                     0.19 & 0.04 & 0.13 &  0.756 & 45.32 & 34.38 & 70.01
                    \\ \hline
        
        Alpaca* & 
                0.18 & 0.04 & 0.13 & 0.721 & 35.19 & 27.82 & 52.80 &
                0.17 & 0.03 & 0.13 &  0.740 & 21.78 & 15.06 & 32.29
                \\ \hline
        Falcon-7b-inst* &  
                    0.21 & 0.05 & 0.15 & 0.739 & 33.96 & 27.82 & 52.63 &
                    0.22 & 0.05 & 0.15 & 0.761 & 24.99 & 18.10 & 38.80
                    \\ \hline  
        Flan-T5-xxl* &  
                0.45 & 0.20 & 0.36 & 0.851 & \textbf{17.77} & 9.69 & \textbf{24.08} &
                0.32 & 0.10 & 0.25 & 0.794 & \textbf{12.82} & 6.92 & 18.10
                \\ \hline   \hline
        Dolly* &  
                 0.17 & 0.03 & 0.12 & 0.723 & 57.42 & 42.03 & 80.99 &
                 0.18 & 0.03 & 0.13 & 0.755 & 44.52 & 30.14 & 62.91
                \\ \hline  
        Dolly** &  
               0.12 & 0.01 & 0.09 & 0.727 & 53.59 & 43.41 & 79.20 &
               0.14 & 0.02 & 0.10 & 0.740 & 51.06 & 35.95 & 70.97    
               \\ \hline \hline
               
        Flan-T5-small* &   
                  0.42 & 0.19 & 0.33 & 0.844 & 24.38 & 8.71 & 27.25 &
                  0.22 & 0.07 & 0.17 & 0.742 & 17.11 & 5.54 & \textbf{15.62}
                  
                  \\ \hline
        Flan-T5-small** &   
                   0.43 & 0.20 & 0.33 & 0.845 & 24.45 & \textbf{8.70} & 27.33 &
                   0.23 & 0.07 & 0.18 & 0.741 & 17.16 & 5.63 & 15.86
                   \\ \hline \hline
        Flan-T5-base* &                  
                 0.46 & 0.23 & 0.36 & 0.856 & 24.29 & 9.14 & 28.01 &
                 0.31 & 0.11 & 0.24 & 0.785 & 18.82 & \textbf{4.80} & 15.93
                 \\ \hline
        Flan-T5-base** &                  
                0.47 & 0.23 & 0.37 & 0.857 & 24.23 & 9.09 & 27.92 &
                0.32 & 0.12 & 0.25 & 0.787 & 18.58 & 4.82 & 15.97
                 \\ \hline \hline
                 
        Flan-T5-large* &  
                  0.44 & 0.19 & 0.35 & 0.849 & 18.38 & 10.24 & 25.25 &
                  0.29 & 0.09 & 0.23 & 0.793 & 13.23 & 7.59 & 18.75
                  \\ \hline
         Flan-T5-large** &  
                 0.46 & 0.20 & 0.36 & 0.851 & 18.38 & 10.24 & 24.22 &
                 0.30 & 0.10 & 0.24 & 0.793 & 13.24 & 7.52 & 18.72 
                  \\ \hline \hline
                      
        Flan-T5-xl* &  
               0.48 & 0.26 & 0.39 & 0.862 & 24.37 & 9.87 & 28.21 &
               0.37 & 0.16 & 0.29 & 0.824 & 18.10 & 5.71 & 18.13
               \\ \hline 
        Flan-T5-xl** &  
               0.49 & 0.26 & 0.40 & 0.863 & 24.41 & 9.92 & 28.27 &
               0.38 & 0.17 & 0.31 & 0.827 & 18.17 & 5.71 & 18.13 
               \\ \hline \hline

        Llama-7b-chat* &  
                       0.18 & 0.04 & 0.13 & 0.721 & 32.41 & 25.65 & 48.58 &
                       0.21 & 0.05 & 0.15 &  0.765 & 31.46 & 22.54 & 47.39
                       \\ \hline 
        Llama-7b-chat**$^\pm$ &  
               0.18 & 0.05 & 0.14 & 0.721 & 31.66 & 25.03 & 47.45 &
               0.21 & 0.06 & 0.15 & 0.760 & 30.54 & 22.17 & 46.17 
               \\ \hline  \hline
        T0pp-11b** &  
               0.46 & 0.20 & 0.36 & 0.853 & 18.91 & 10.15 & 25.69 &
               0.33 & 0.11 & 0.26 & 0.804 & 15.19 & 7.58 & 20.45
               \\ \hline  \hline
        GPT-3.5 turbo &  
               0.14 & 0.06 & 0.10 & 0.746 & 189.27 & 80.86 & 195.89 &
               0.14 & 0.05 & 0.11 & 0.741 & 140.31 & 62.33 & 150.97                
               \\ \hline  \hline
    \end{tabular}
    \end{adjustbox}
    \caption{Zero-shot evaluation of models on conversational summarization task on all the models (including those which were found to be not suitable for conversational summarization). This table is a superset of Table~\ref{tab:results}. The * indicates the models are instruction tuned. ** Indicates that the models were further instructed tuned for conversational summarization task. $^\pm$ indicates that the model was trained using adapters.}
    \label{tab:results_all}
\end{table*}

\begin{table}[]
    \centering
    \begin{adjustbox}{width=\textwidth}
    \begin{tabular}{| l   
                    | c | c | c 
                    | c | c | c 
                    | c | c | c 
                    | c | c | c |}
    \hline
           \multirow{1}{*}{Model}  
           & \multicolumn{3}{c|}{Samsum}  
           & \multicolumn{3}{c|}{Samsum - FT}
           & \multicolumn{3}{c|}{dialogsum}
           & \multicolumn{3}{c|}{dialogsum -FT} \\ \cline{2-13}
            & R1 & RL & B-Score 
            & R1 & RL & B-Score 
            & R1 & RL & B-Score 
            & R1 & RL & B-Score \\ \hline
            Flan-T5-small 
            & 0.42 & 0.33 & 0.846 
            & 0.43 & 0.34 & 0.862
            & 0.24 & 0.20 & 0.751
            & 0.24 & 0.20 & 0.753
            \\ \hline
            Flan-T5-base & 0.46 & 0.37 & 0.857 
            & 0.47 & 0.38 & 0.857
            & 0.33 & 0.26 & 0.793
            & 0.34 & 0.26 & 0.794
            \\ \hline
            Flan-T5-large & 0.45 & 0.36 & 0.852 
            & 0.49 & 0.40 & 0.886
            & 0.31 & 0.25 & 0.802
            & 0.30 & 0.25 & 0.801
            \\ \hline
            Flan-T5-xl & 0.49 & 0.39 & 0.863 
            & 0.49 & 0.40 & 0.863 
            & 0.39 & 0.31 & 0.824
            & 0.39 & 0.32 & 0.836
            \\ \hline
            Llama-7b-chat* & 0.32 & 0.24 & 0.808 
            & 0.33 & 0.24 & 0.809
            & 0.26 & 0.21 & 0.789
            & 0.29 & 0.21 & 0.792
            \\ \hline
            GPT-3.5 & 0.21 & 0.14 & 0.748 
            & - & - & -
            & 0.24 & 0.20 & 0.775 
            & - & - & -
            \\ \hline
    \end{tabular}
    \end{adjustbox}
    \caption{Average best performing prompt result. }
    \label{tab:results_finetuning}
\end{table}

\subsection{Dataset resources}

1. Dialogsum: \url{https://huggingface.co/datasets/knkarthick/dialogsum} 

2. Samsum: \url{https://huggingface.co/datasets/samsum}

3. Promptsource: \url{https://github.com/bigscience-workshop/promptsource/tree/main/promptsource/templates/samsum}

\subsection{Model hyper-parameters}

We num train epochs=25, train batch size=2, warmup steps=500,   weight decay=0.01, learning rate=5e-10, AdamW optimizer for Flan-T5 models and Dolly models. We used FP32 for Flan-T5 and Fp16 precision for Dolly model weights. 

For Lama-7b-chat finetuning we used LORA \cite{hu2021lora} adapters for finetuning. 

lora\_r = 8 \\ 
lora\_alpha = 16 \\
lora\_dropout = 0.05 \\
lora\_target\_modules= [ \\
    ``q\_proj", \\
    ``v\_proj", \\
]
with , \\
num\_train\_epochs=10, \\
per\_device\_train\_batch\_size=2, \\ 
per\_device\_eval\_batch\_size=2,  \\ 
warmup\_steps=500, \\ 
weight\_decay=0.01,\\ 

When inferencing with GPT3.5 we used temperature=0.2, max tokens=500, top p=1, frequency penalty=0, presence penalty=0. 

\begin{figure*}
    \centering
    \includegraphics[width=1.0\textwidth]{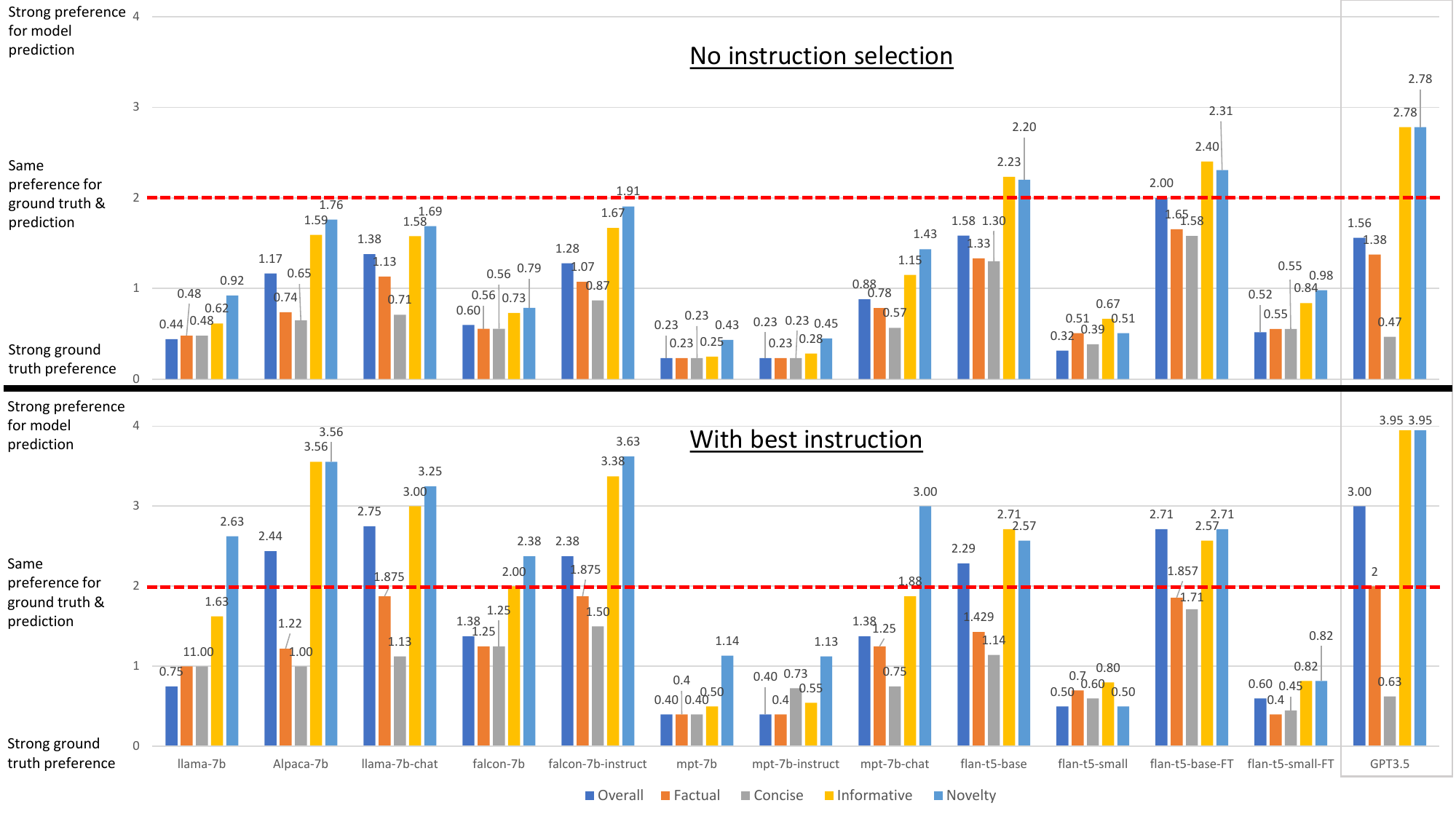}
    \caption{Shows human evaluation of summaries generated by the models}
    \label{fig:human-ratings}
\end{figure*}

\begin{table}[]
    \centering
    \begin{tabular}{c|c|c}
    \hline
         1 & Generate a summary for this dialogue & Used\\
         2 & Summarize: & Used\\
         3 & To sum up this dialog & Used\\
         4 & Given the above dialogue write a summary & Used\\
         5 & Summarize this dialogue: & Used\\
         6 & Sum up the following dialogue & Used\\ 
         7 & Write a dialogue that match this summary & Not Used\\ \hline
    \end{tabular}
    \caption{Prompts from the Promptsource dataset and if they're used for generating the summary in this work. }
    \label{tab:prompts}
\end{table}


\begin{table*}[]
    \centering
    \begin{adjustbox}{width=\textwidth}
    \begin{tabular}{| l   
                    | c | c | c | c | c |
                    | c | c | c | c | c | }
    \hline
            \multirow{1}{*}{Model} 
                & \multicolumn{5}{c|}{Samsum} 
                & \multicolumn{5}{c|}{Dialogsum}
                \\ \cline{2-11}
                & R1(p/r) & R2(p/r) & RL(p/r) & N1 & N2 
                & R1(p/r) & R2(p/r) & RL(p/r) & N1 & N2
                \\ \hline
         Flan-T5-small* &   
                   0.72/0.30 & 0.37/0.15 & 0.58/0.24 & 9.15 & 28.41 &
                   0.80/0.20 & 0.50/0.12 & 0.63/0.16 & 5.75 & 16.10
                  
                  \\ \hline
         Flan-T5-base* &                  
                   0.74/0.31 & 0.37/0.15 & 0.59/0.25 & 9.60 & 29.21 &
                   0.84/0.23 & 0.55/0.15 & 0.70/0.19 & 4.97 & 16.41
                 \\ \hline
         Flan-T5-large* &  
                   0.68/0.20 & 0.24/0.07 & 0.51/0.15 & 10.75 & 26.34 &
                   0.70/0.13 & 0.28/0.05 & 0.53/0.10 & 7.86 & 19.32
                  \\ \hline        
        Llama-7b &  
                   0.39/0.21 & 0.10/0.05 & 0.28/0.14 & 49.66 & 89.49 &
                   0.43/0.21 & 0.11/0.05 & 0.29/0.13 & 23.49 & 48.84
                  \\ \hline
        Falcon-7b &  
                   0.34/0.27 & 0.07/0.17 & 0.20/0.17 & 51.32 & 93.59 &
                   0.46/0.21 & 0.11/0.05 & 0.29/0.12 & 27.64 & 56.09
                   \\ \hline
        MPT-7b &  
                  0.34/0.07 & 0.10/0.07 & 0.22/0.17 & 47.53 & 86.49 &
                  0.42/0.24 & 0.11/0.07 & 0.25/0.14 & 36.22 & 70.24
                \\ \hline
        Stablelm-7b &  
                  0.31/0.24 & 0.06/0.05 & 0.19/0.15 & 48.30 & 87.20 &
                  0.42/0.22 & 0.11/0.06 & 0.25/0.13 & 34.87 & 68.43
                         \\ \hline
        MPT-7b-inst* &  
                  0.31/0.26 & 0.08/0.07 & 0.19/0.07 & 51.96 & 92.31 &
                  0.54/0.18 & 0.14/0.05 & 0.37/0.12 & 41.91 & 76.98
                         \\ \hline
        Dolly* &  
                 0.38/0.30 & 0.10/0.08 & 0.24/0.19 & 44.26 & 84.53 &
                 0.49/0.27 & 0.16/0.09 & 0.29/0.16 & 31.35 & 64.89
                \\ \hline   
        Alpaca* & 
                0.39/0.21 & 0.10/0.05 & 0.28/0.14 & 29.33 & 55.03 &
                0.47/0.20 & 0.14/0.06 & 0.34/0.14 & 15.61 & 33.22
                \\ \hline
        Falcon-7b-inst* &  
                 0.41/0.20 & 0.08/0.05 & 0.28/0.13 & 29.27 & 54.88 &
                 0.54/0.18 & 0.14/0.05 & 0.37/0.05 & 18.81 & 40.02
                    \\ \hline  
        Flan-T5-xl* &  
                 0.73/0.31 & 0.37/0.15 & 0.57/0.24 & 10.36 & 29.42 &
                 0.82/0.22 & 0.48/0.13 & 0.67/0.18 & 5.90 & 18.68
               \\ \hline
        Flan-T5-xxl* &  
                  0.69/0.19 & 0.25/0.07 & 0.52/0.15 & 10.18 & 25.11 &
                  0.74/0.14 & 0.32/0.06 & 0.57/0.11 & 7.01 & 18.54
                \\ \hline
        MPT-7b-chat* & 
                 0.32/0.29 & 0.09/0.08 & 0.20/0.19 & 52.65 & 96.40 &
                 0.45/0.27 & 0.13/0.08 & 0.26/0.16 & 35.80 & 72.21
                    \\ \hline
        Llama-7b-chat* &  
                 0.38/0.21 & 0.09/0.21 & 0.28/0.06 & 27.05 & 50.68 &
                 0.50/0.14 & 0.14/0.06 & 0.33/0.14 & 23.49 & 48.84
                       \\ \hline
        Flan-T5-small** & 
                 0.72/0.30 & 0.38/0.15 & 0.58/0.25 & 9.13 & 28.51 &
                 0.80/0.20 & 0.49/0.12 & 0.62/0.16 & 5.83 & 16.35
                \\ \hline
       Flan-T5-base** &  
                 0.74/0.30 & 0.38/0.15 & 0.58/0.25 & 9.55 & 29.12 &
                 0.84/0.23 & 0.54/0.15 & 0.69/0.19 & 4.99 & 16.45                  
                    \\ \hline  
        Flan-T5-large** &  
                 0.68/0.20 & 0.24/0.07 & 0.51/0.15 & 10.77 & 26.36 &
                 0.70/0.13 & 0.28/0.05 & 0.53/0.10 & 7.79  & 19.29
               \\ \hline
        Flan-T5-xl** &  
                 0.73/0.31 & 0.37/0.15 & 0.57/0.24 & 10.41 & 29.48 &
                 0.82/0.22 & 0.48/0.13 & 0.67/0.13 & 5.90 & 18.67
                \\ \hline
        Dolly ** & 
                   0.31/0.22 & 0.06/0.04 & 0.22/0.13 & 45.58 & 82.58 &
                   0.43/0.23 & 0.08/0.05 & 0.24/0.13 & 33.69 & 67.67
                    \\ \hline
        T0pp-11b ** & 
                   0.69/0.25 & 0.25/0.08 & 0.52/0.16 & 10.65 & 26.80 &
                   0.74/0.16 & 0.32/0.07 & 0.56/0.12 & 7.84 & 21.06
                    \\ \hline
        Llama-7b-chat**$^\pm$ &  
                 0.38/0.21 & 0.09/0.06 & 0.28/0.14 & 26.34 & 49.43 &
                 0.50/0.13 & 0.13/0.06 & 0.33/0.13 & 22.94 & 47.60
                       \\ \hline

    \end{tabular}
    \end{adjustbox}
    \caption{Table shows the Rouge and Novel 1-gram and Novel 2-gram scores of the generated summary compared to the . High Rouge   of copying mechanisms in the summary generated}
    \label{tab:copying_p}
\end{table*}

\begin{figure}
  \begin{center}
    \includegraphics[width=0.95\textwidth]{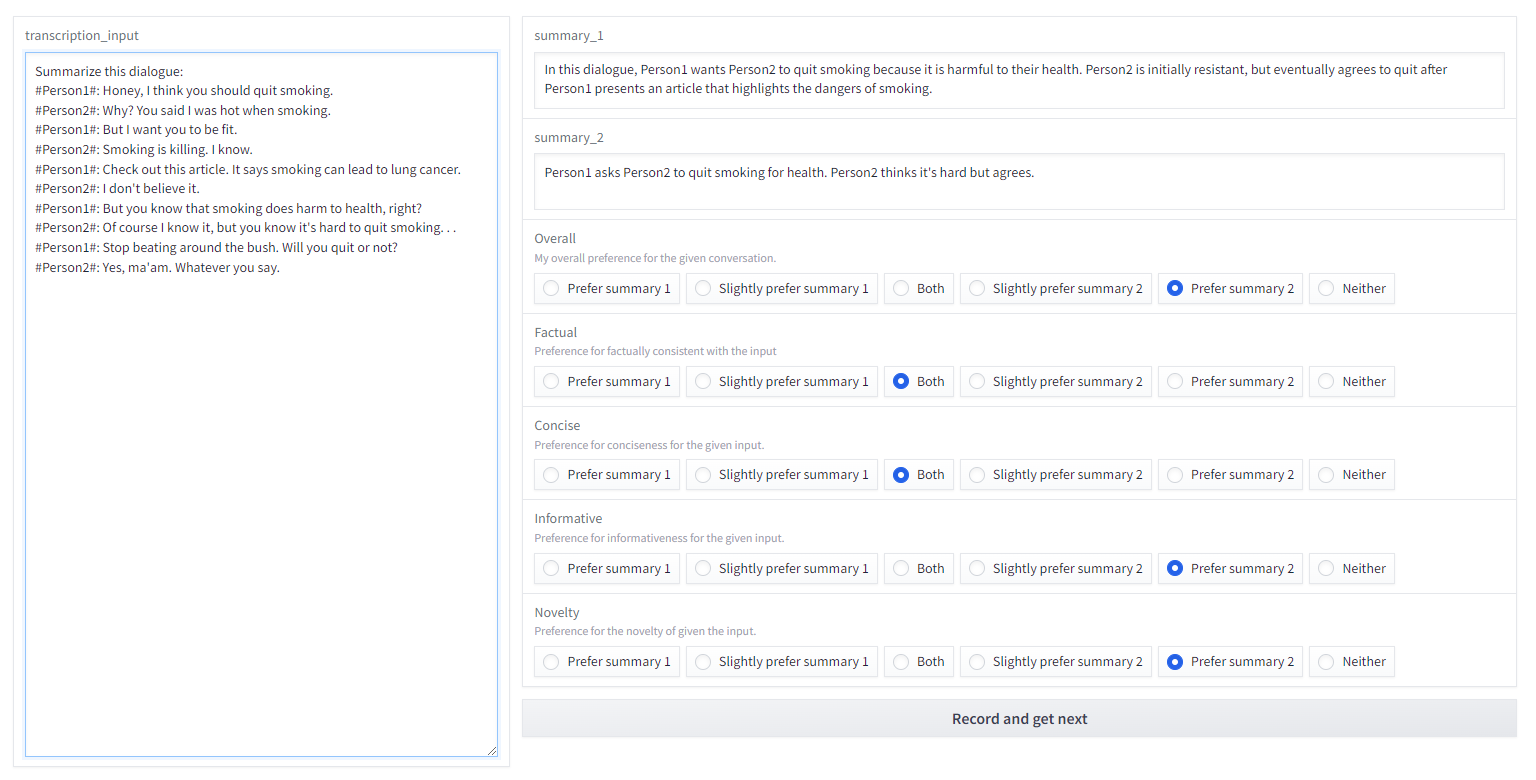}
  \end{center}
  \label{fig:ratings_ui}
  \caption{Interface for humans rating the summaries comparing two summaries. }
\end{figure}

\begin{figure*}
    \centering
    \includegraphics[width=1.0\textwidth]{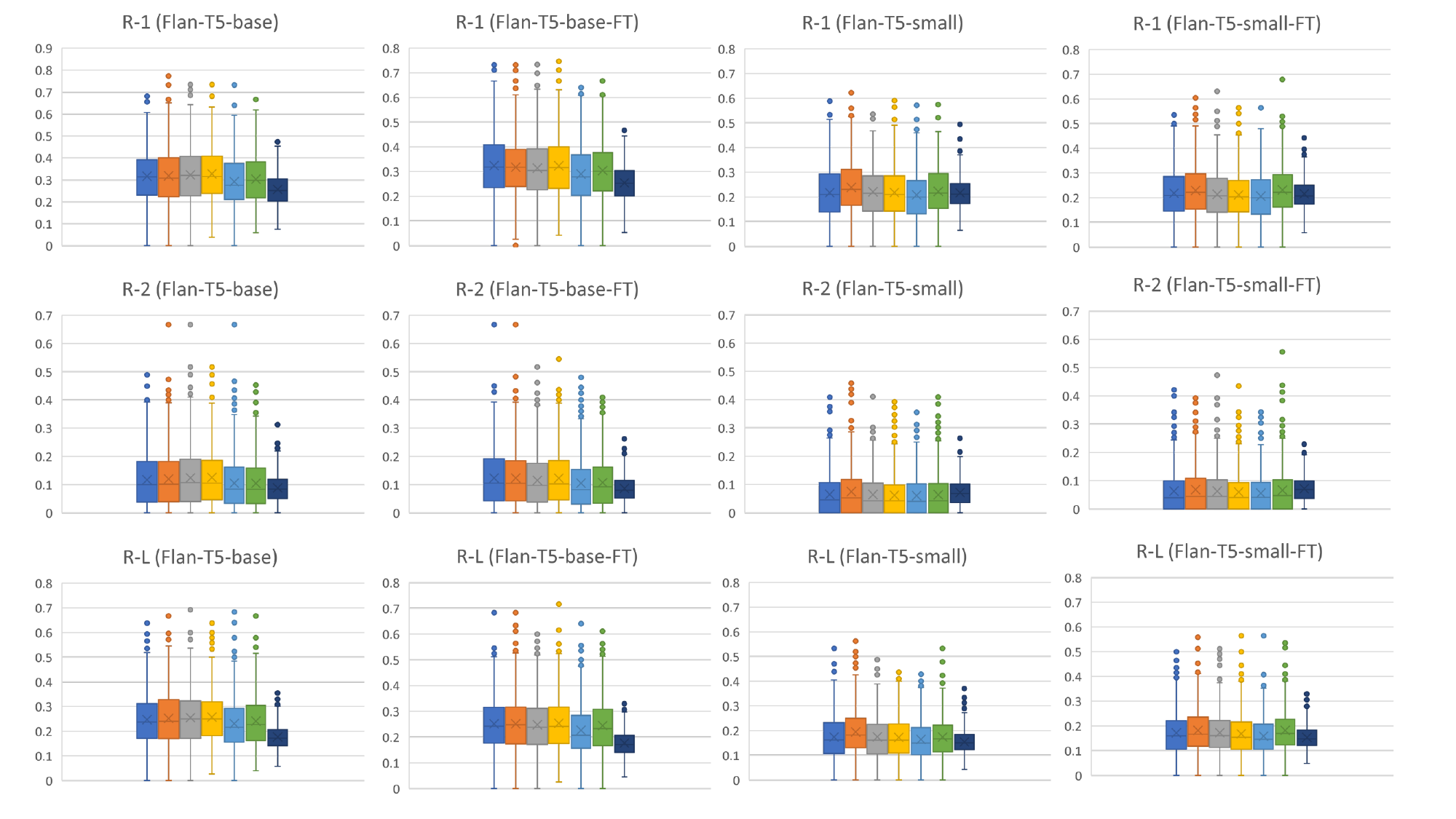}
    \caption{Box plots with Rouge scores per prompt from Prompsource dataset and various LLMs. Flan models (instruction tined) models seem to be more robust to different prompts. ROUGE scores for all the prompts in Table~\ref{tab:prompts}. }
    \label{fig:enter-label}
\end{figure*}

\begin{figure*}
    \centering
    \includegraphics[width=1.0\textwidth]{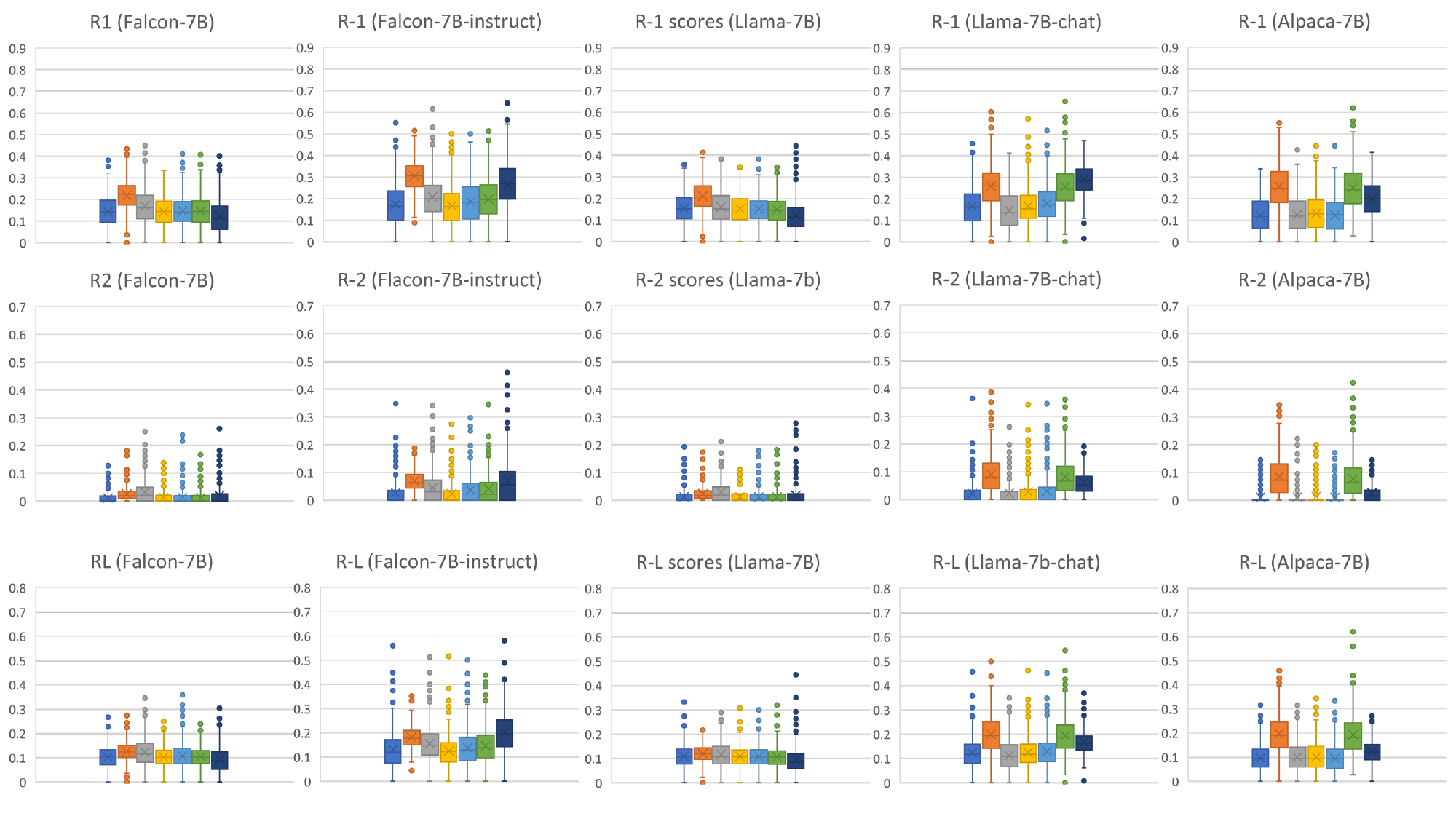}
    \caption{Box plots with Rouge scores per prompt from Prompsource dataset for various LLMs. Instruction tunings seems to improve the ROUGE scores across prompts, however the performance on different prompts seems to be an issue. ROUGE scores for all the prompts in Table~\ref{tab:prompts}.}
    \label{fig:enter-label}
\end{figure*}

\begin{table*}
    \centering
    \caption{Example (not handpicked). Chosen randomly from the dataset. }
    \begin{adjustbox}{width=\textwidth}
    \begin{tabular}{| c | c | }
    \hline
        (Conversation) & (Ground truth) \\ 
         \makecell[l]{
             Person1: Hello, how are you doing today? \\
             Person2: I ' Ve been having trouble breathing lately. \\
             Person1: Have you had any type of cold lately? \\ Person2: No, I haven ' t had a cold. \\
             \hspace{5mm} I just have a heavy feeling in my chest \\
             \hspace{5mm} when I try to breathe. \\ 
             Person1: Do you have any allergies that you know of? \\ 
             Person2: No, I don ' t have any allergies that I know of. \\
             Person1: Does this happen all the time or mostly \\
             \hspace{5mm} when you are active? \\
             Person2: It happens a lot when I work out. \\
             Person1: I am going to send you to a pulmonary \\ \hspace{5mm} specialist who can run tests on you for asthma. \\
             Person2: Thank you for your help, doctor.
            }  & 
         \makecell{Person2 has trouble breathing. \\
         The doctor asks Person2 about it \\ 
         and will send Person2 to a pulmonary \\
         specialist.
            }  \\ \hline
          \makecell{(Flan-T5-small) \\
          I am going to send you to a pulmonary specialist\\
          who can run tests on you for asthma. \\
          Do you have any allergies that you know?
          }
          &  \makecell{(Flan-T5-base)\\
          Person1 is going to send Person2 to a pulmonary \\ 
          specialist who can run tests on Person2 for asthma.\\
          He will send Person1 and Person2 a message.
          }  \\ \hline
          \makecell{(Flan-T5-small**) \\
          Person2: Hi, how are you doing today?\\
          I 've been having problems breathing lately.\\
          I haven 't had a cold. I just have a heavy \\
          feeling in my chest when I try to breathe\\
          Person1: Do you have any allergies that you know?
          }
          &  \makecell{(Flan-T5-base**) \\
          Person2 has been having trouble breathing lately. \\
          She has a heavy feeling in her chest when she try \\
          to breathe. Person1 is going to send her to a \\
          pulmonary specialist who can run tests on her \\
          for asthma.
          }  \\ \hline
          \makecell{(Llama-7b) \\
          1. Describe your dialogue: \\
          2. How should you greet the patient? (1 Marks) \\
          3. What should you ask about the lifestyle? (1 Marks)\\
          4. What symptoms would you suspect? (1 Marks) \\
          5. List three diagnostic tests that your friend \\
          doctor would run on a patient suffering from your \\ partner patient disease. (2 Marks) \\
          }
          &  \makecell{(Alpaca-7b) \\
          The patient has been having difficulty \\
          breathing and it is a recurring issue. \\ 
          The doctor notices that it happens mostly \\
          when the patient is active, so they suggest \\
          sending the patient to a pulmonary specialist\\
          for tests to rule out asthma. The patient is \\
          thankful for the doctor's help.
          }  \\ \hline
          \makecell{(Llama-7b-chat) \\
          Person 1 (doctor) is interacting with Person 2 (patient)\\
          during a consultation. Person 2 has been \\
          experiencing difficulty breathing, and the doctor\\
          asks questions to gather more information and \\
          determine the cause. The doctor suspects asthma \\
          and recommends a pulmonary specialist for further \\
          tests.
          }
          &  \makecell{(Llama-7b-chat**) \\
          Summary: The two people in the dialogue are \\
          discussing symptoms related to breathing difficulties.\\
          Person two explains that they have been experiencing a\\
          heavy feeling in their chest when trying to breathe, \\
          but have not had any colds or allergies. The first \\
          person suggests that they may have asthma and sends\\
          them to a pulmonary specialist for further testing.
          }  \\ \hline
          \makecell{(Falcon-7b) \\
          Write a summary of Person1: \\ 
          ‘Person1 has a heavy feeling in the chest\\
          when he/she tries to breathe.’ Person1 goes\\
          to the primary care physician who diagnoses \\
          him/her with a lung infections. Person1: \\
          and the doctor sends him/her to another doctor\\
          for further tests. The doctor has been breathing\\
          difficulties. The doctor runs tests like a lung\\
          disease. A specialist helps Person2: ’s diagnose
          }
          &  \makecell{(Falcon-7b-instruct) \\
          After discussing symptoms with Person1,\\
          an appointment is arranged with a pulmonary \\
          specialist to examine person2 and diagnose any\\
          underlying respiratory issues.
          }  \\ \hline
          \makecell{(Flan-T5-XL) \\
          Person2 has been having trouble breathing lately.\\ Person1 is going to send him to a pulmonary \\
          specialist who can run tests on him for asthma.
          }
          &  \makecell{(Flan-T5-XXL) \\
          Person2 has been having breathing problems lately.\\
          They happen primarily when he exercises, so he \\
          might have Asthma.
          }  \\ \hline
    \end{tabular}
    \end{adjustbox}
    \label{tab:example_main}
\end{table*}

    \centering
    \begin{tabular}{| c | c | }
    \hline
          \makecell{(Flan-T5-Large)\\
          Person2 has trouble breathing.\\
          She has not had any cold, and her \\
          chest is heavy when she tries to breath.\\
          At Person1 she will have tests for asthma.
          }
          &  \makecell{(Flan-T5-Large**)\\
          Person2 has trouble breathing because of \\
          allergies. She may have asthma and a \\
          specialist will run tests on her.
          }  \\ \hline
          \makecell{(MPT-7b) \\
          Person2: I been having difficulty in breathing\\
          lately.Person2: Never, I have just a heavy \\
          feeling in my chest when I try to breathe.\\
          Person2: I have no allergies that I \\
          know of.The D.P.R.I.V.E.R.S.\\
          Copyright© \\
            \begin{CJK*}{UTF8}{gbsn}
            コスメを楽しくできるガウ！, 
            \end{CJK*}
            \\
            2016 All Rights \\Reserved.We're all set for our launch!
          }
          &  \makecell{(MPT-7b-chat) \\
          Person1 and Person2 are speaking with each \\
          other. Person1 is listening to Person2's \\
          concerns about their breathing difficulties \\
          and suspects that Person2 may have asthma.\\
          Person1 suggests that Person2 should see a\\
          pulmonary specialist and arranges an \\
          appointment for them. Person1 suspects\\
          that Person2 may have asthma because it \\
          happens when Person2 is active which is a\\
          common symptom of asthma. Person1 also\\
          asks about allergies, colds and whether\\
          it happens all the time. 
          }  \\ \hline
          \makecell{(MPT-7b-instruct)\\
          A person that has breathing problems \\
          should consult a doctor that can perform \\
          a pulmonary function test for asthma.\\
          The World Economic Forum has created \\
          a unique and exclusive partnership with \\
          the International Monetary Fund,the World \\
          Bank and the United Nations to tackle\\
          the main issues and challenges of our \\
          time. The U.S. Department of State’s \\ 
          Bureau of East Asian and Pacific Affairs \\
          is seeking young professionals for the \\
          Regional Internship Program (RIP) \\
          at the U.S. Embassy in Beijing.
          }
          &  \makecell{(GPT-3.5)\\
          I appreciate it.  Person 1 is a doctor\\
          who is asking Person 2 how they are \\
          feeling. Person 2 responds that they \\
          have been having trouble breathing lately \\
          and a heavy feeling in their chest. \\
          Person 1 asks if they have had a cold\\
          or any allergies, and Person 2 responds\\
          that they have not. Person 1 asks if this\\
          happens all the time or mostly when they\\
          are active, and Person 2 responds that \\
          it happens a lot when they work out. \\
          Person 1 decides to send Person 2 to a \\
          pulmonary specialist to run tests for asthma.\\
          Person 2 thanks Person 1 for their help.
          }  \\ \hline
          
    \end{tabular}
\begin{figure}
    \centering
    \caption{Rouge, Bert novel n-grams and LR on Dialog sum dataset (contd. in next page)}
    \includegraphics[page=1,scale=0.75]{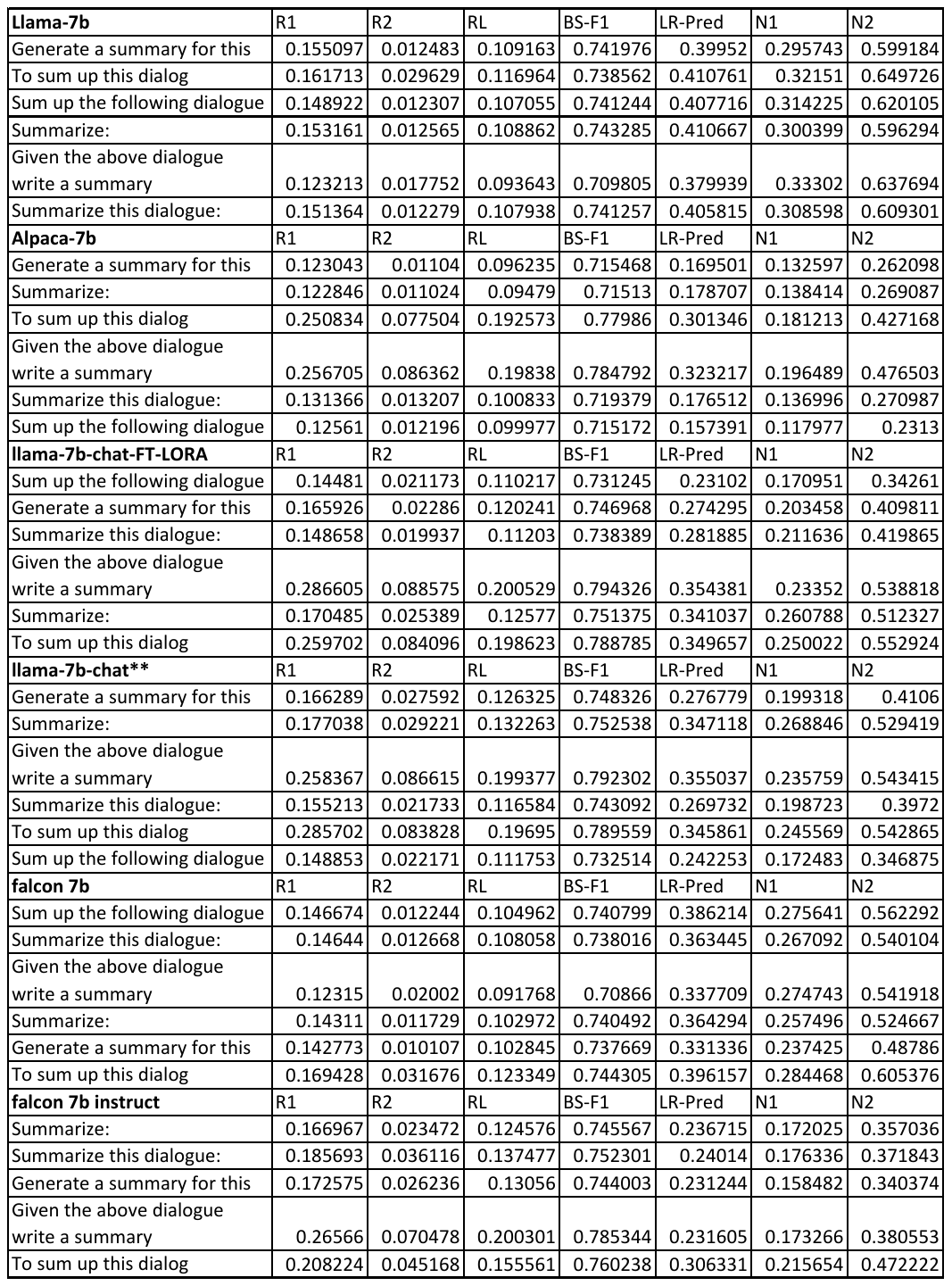}  
    \label{fig:dialsum_prompts_1}
\end{figure}
\begin{figure}
    \centering
     \caption{Rouge, Bert novel n-grams and LR on Dialog sum dataset (contd. in next page)}
    \includegraphics[page=2,scale=0.75]{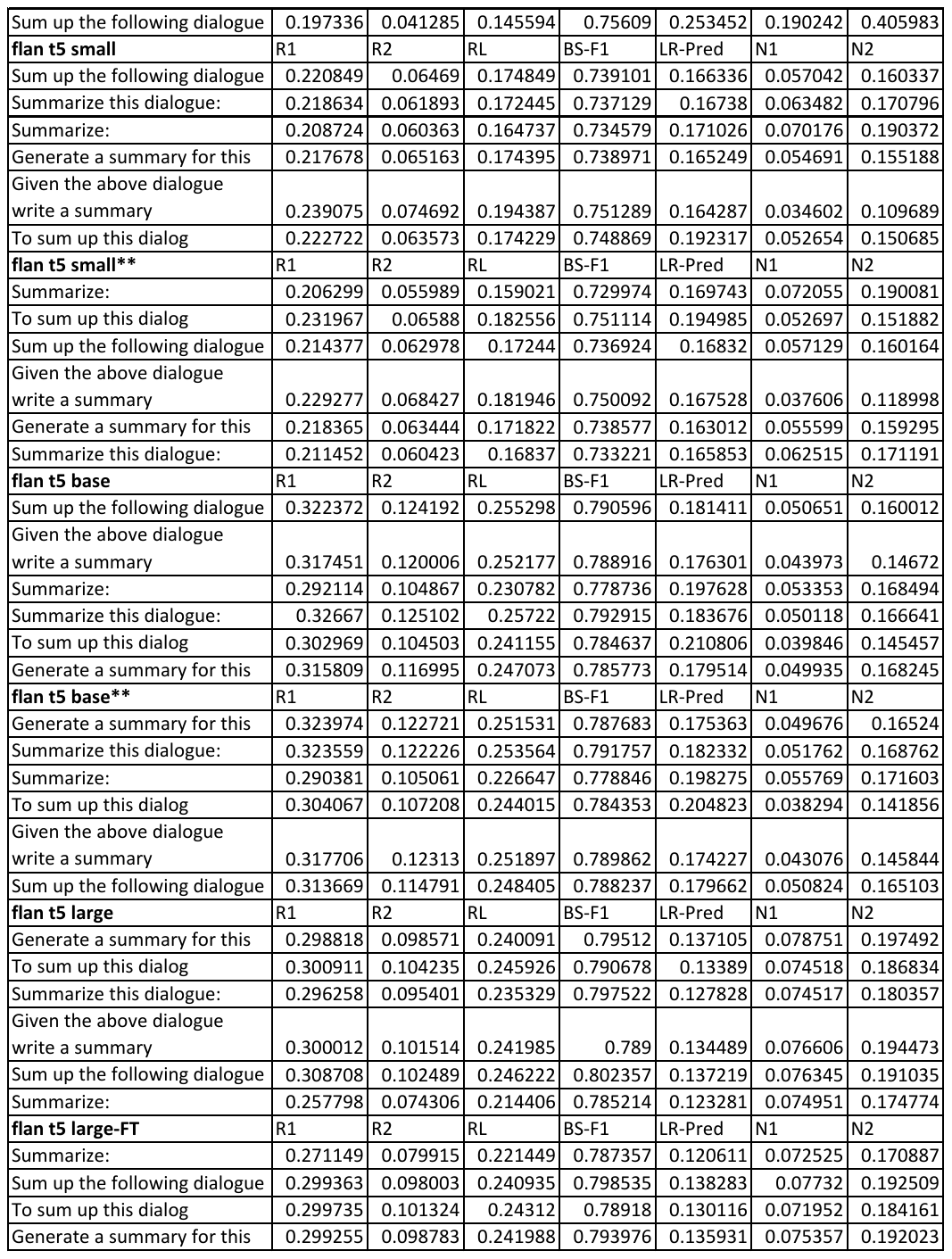}  
   
    \label{fig:dialsum_prompts_2}
\end{figure}
\begin{figure}
    \centering
    \caption{Rouge, Bert novel n-grams and LR on Dialog sum dataset (contd. in next page)}
    \includegraphics[page=3,scale=0.75]{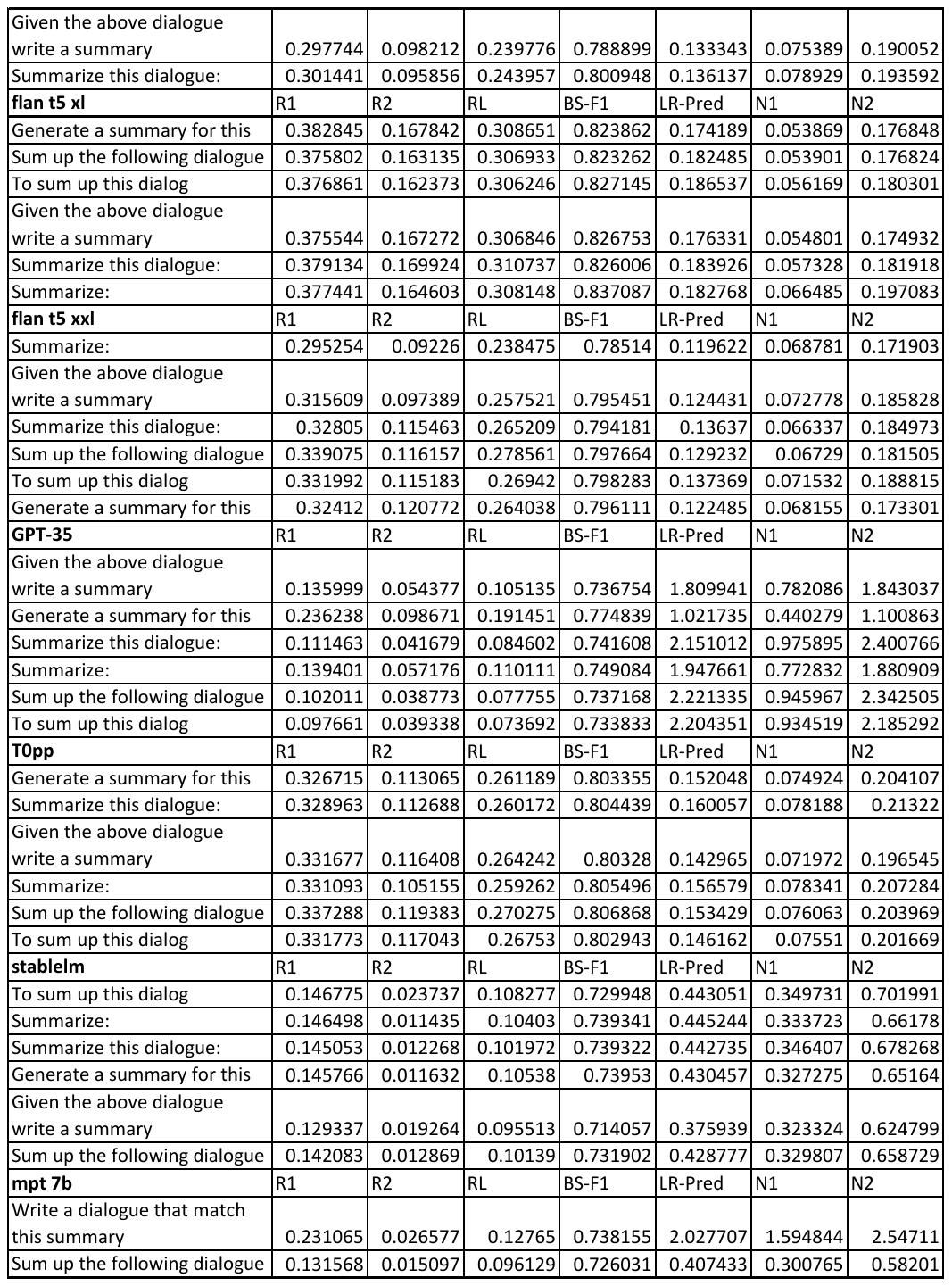}  
    
    \label{fig:dialsum_prompts_3}
\end{figure}
\begin{figure}
    \centering
    \includegraphics[page=4,scale=0.75]{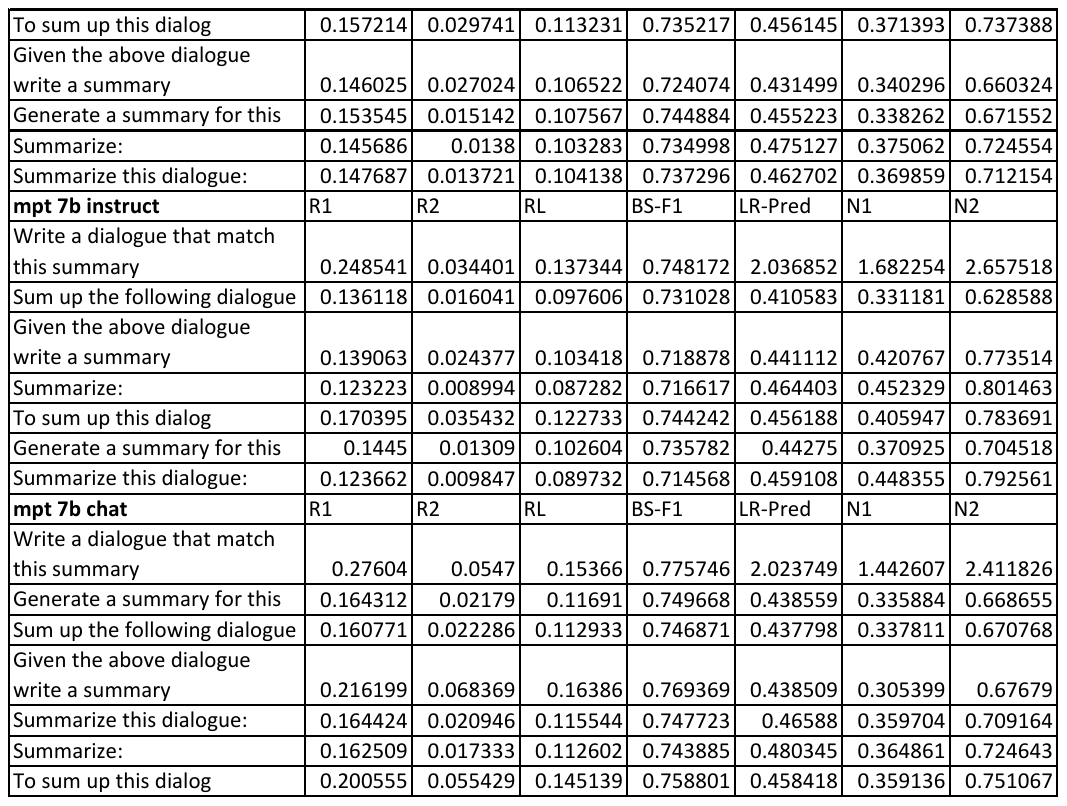}  
    \label{fig:dialsum_prompts_4}
\end{figure}

\begin{figure}
    \centering
    \caption{Rouge, Bert novel n-grams and LR on SAMsum dataset (contd. in next page)}
    \includegraphics[page=1,scale=0.75]{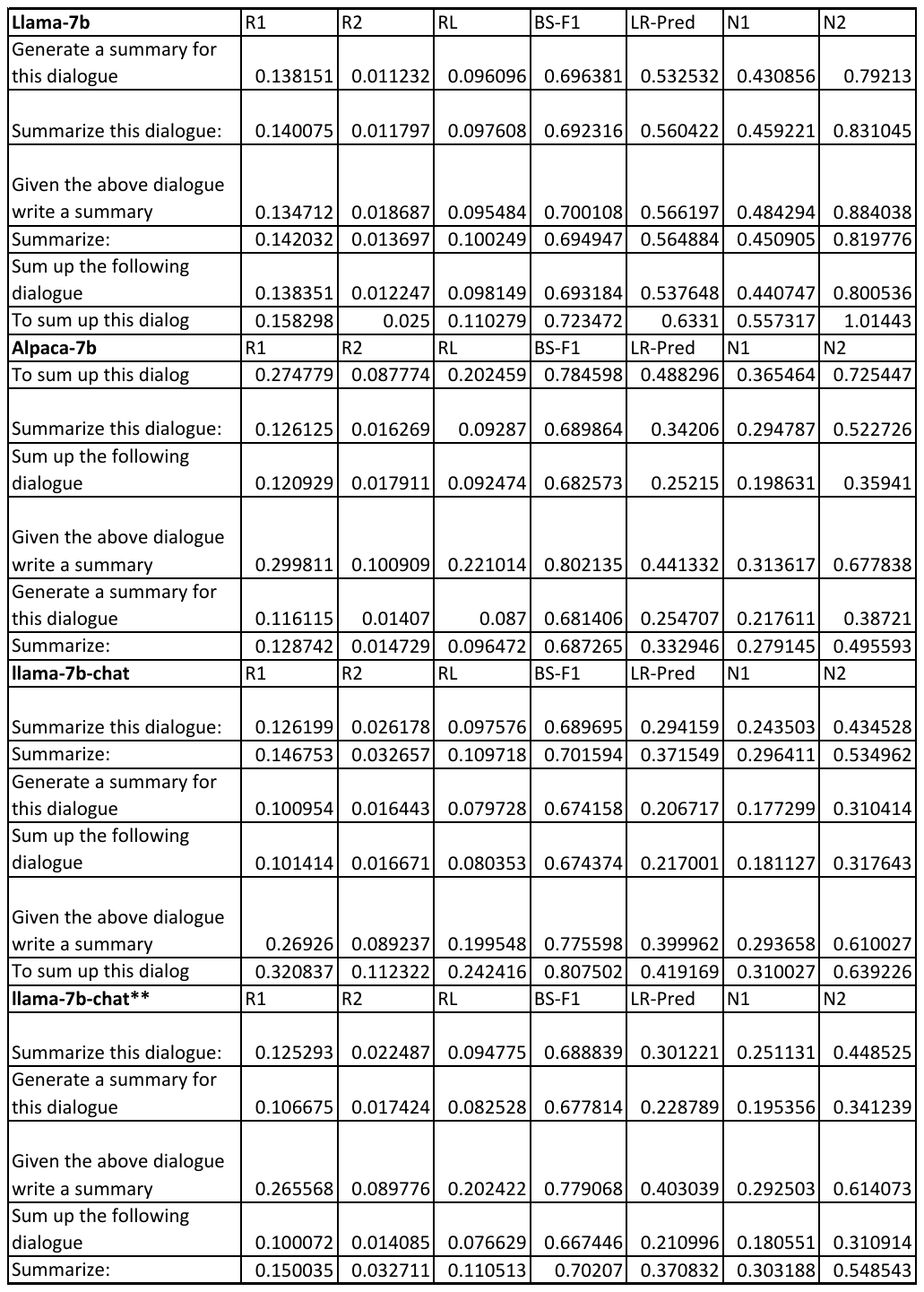}  
    
    \label{fig:samsum_prompts_1}
\end{figure}
\begin{figure}
    \centering
     \caption{Rouge, Bert novel n-grams and LR on SAMsum dataset (contd. in next page)}
    \includegraphics[page=2,scale=0.75]{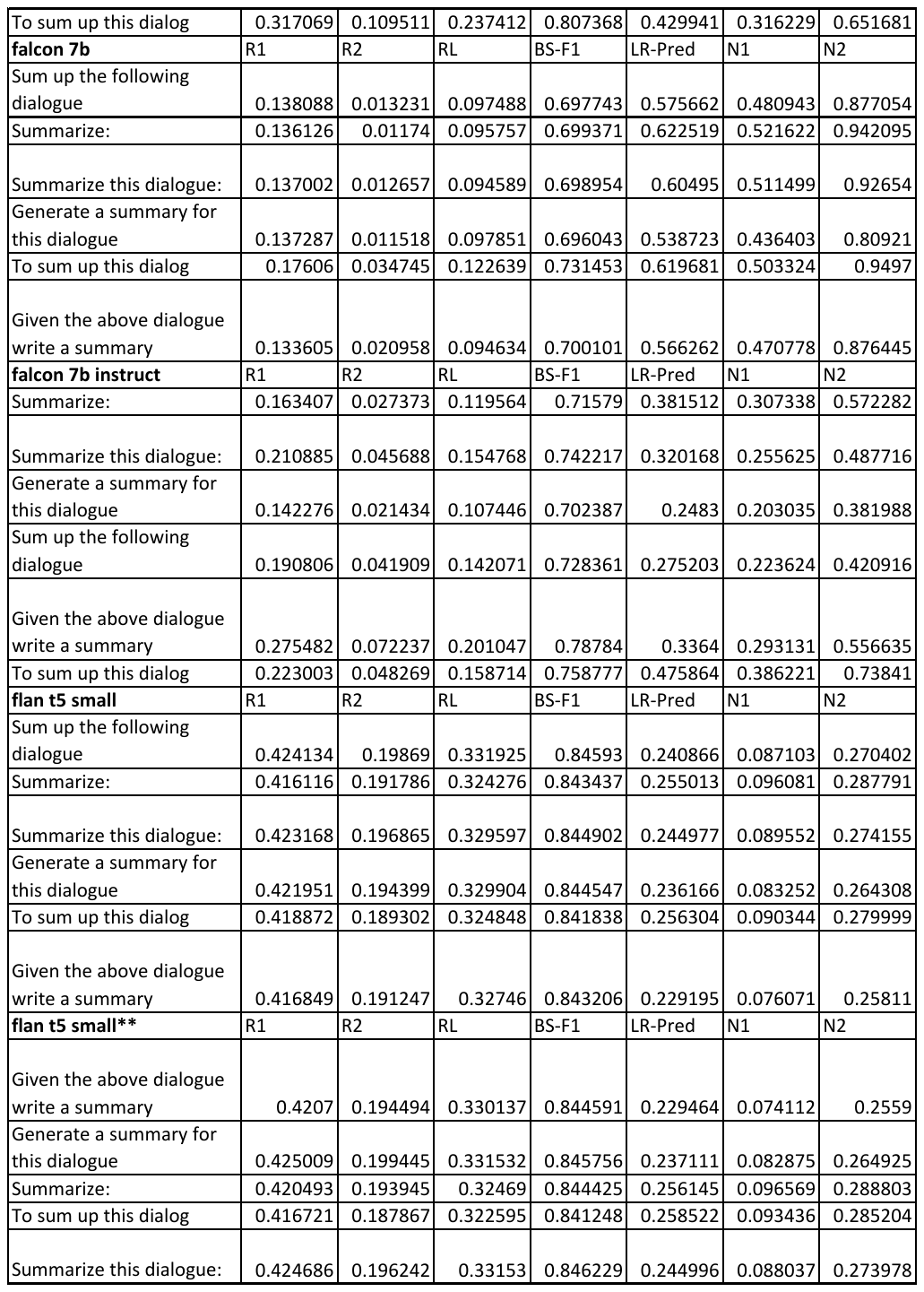}  
   
    \label{fig:samsum_prompts_2}
\end{figure}
\begin{figure}
    \centering
    \caption{Rouge, Bert novel n-grams and LR on SAMsum dataset (contd. in next page)}
    \includegraphics[page=3,scale=0.75]{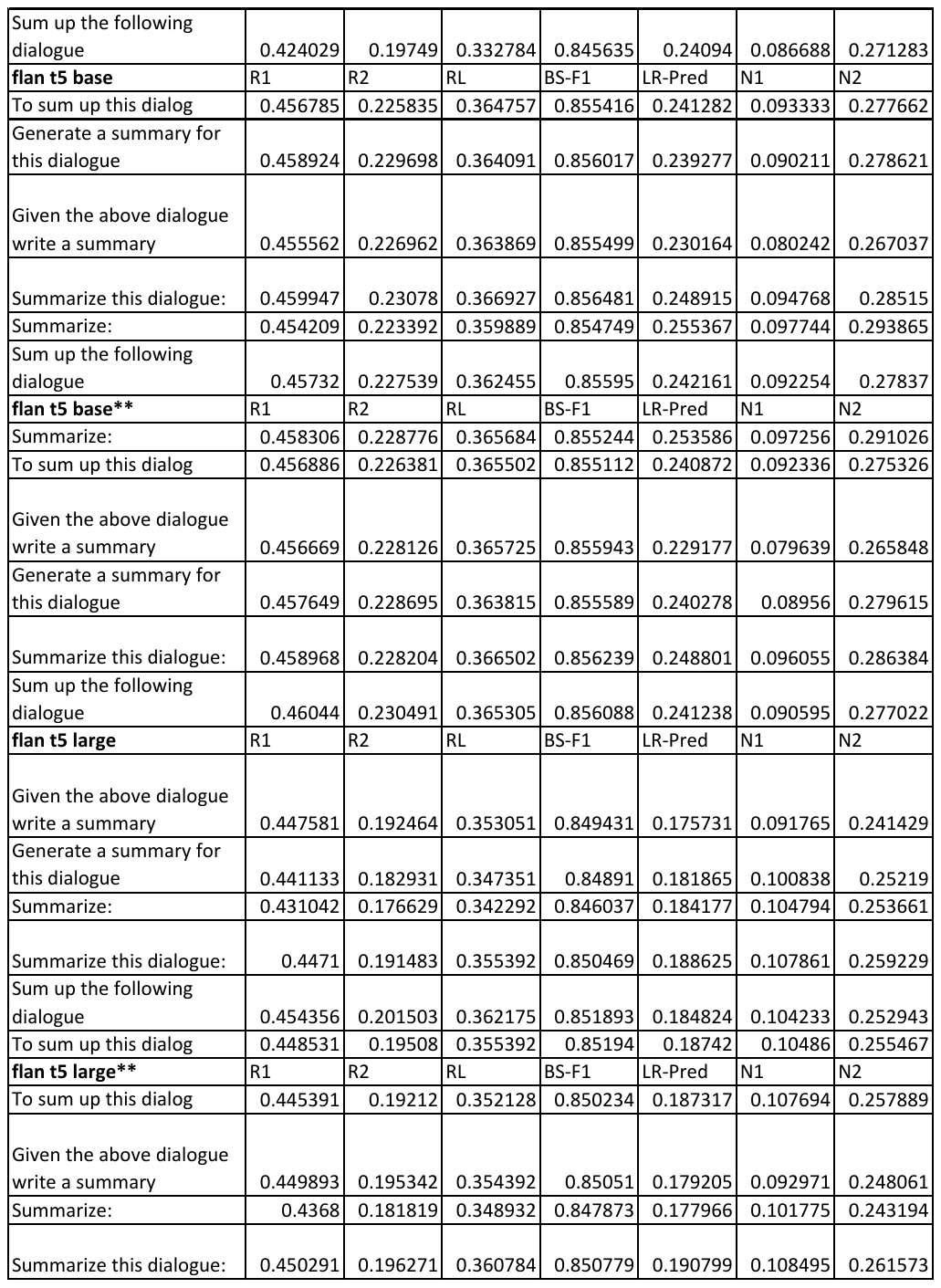}  
    
    \label{fig:samsum_prompts_3}
\end{figure}
\begin{figure}
    \centering
    \caption{Rouge, Bert novel n-grams and LR on SAMsum dataset (contd. in next page)}
    \includegraphics[page=4,scale=0.75]{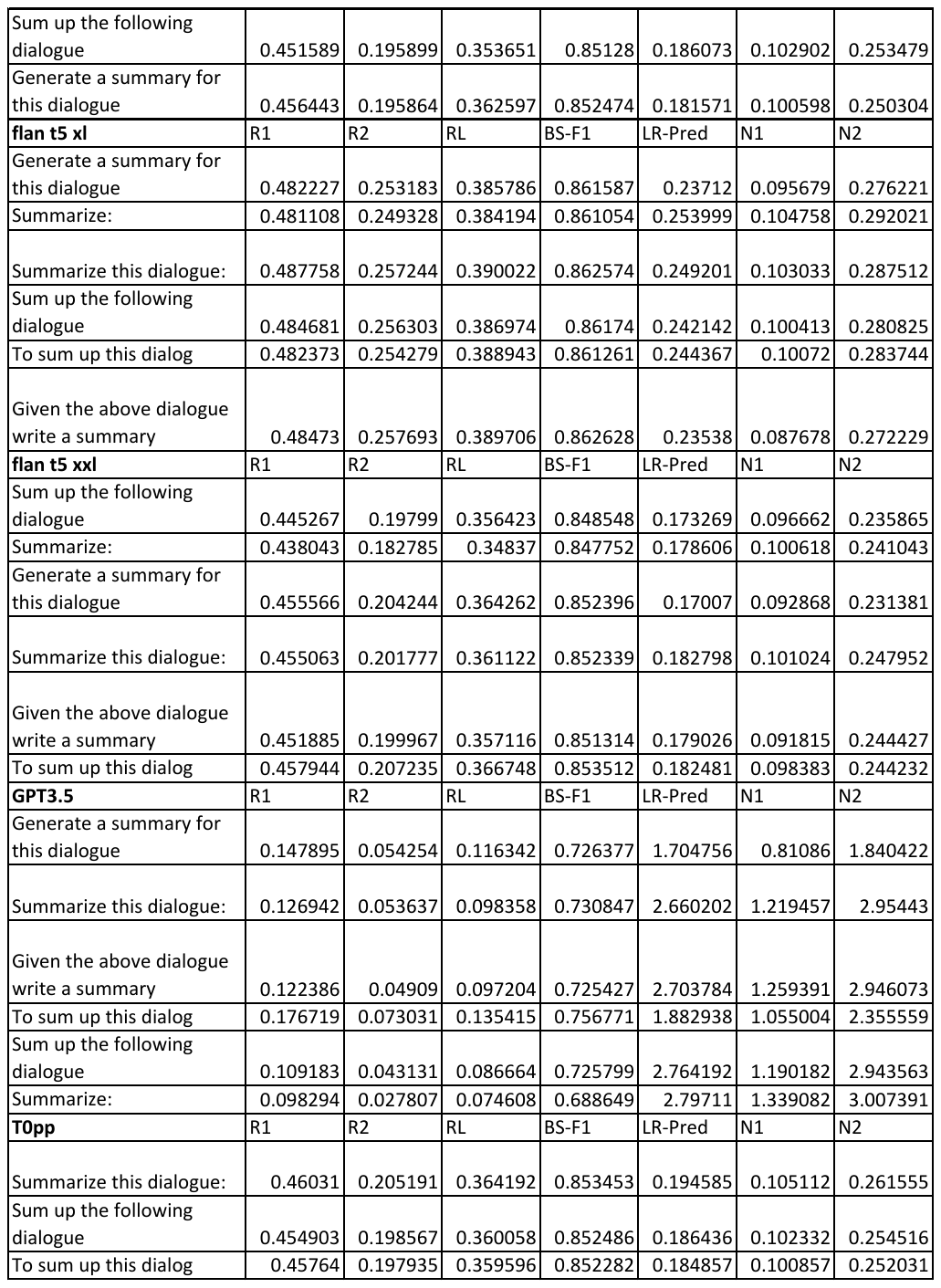}  
     
    \label{fig:samsum_prompts_4}
\end{figure}
\begin{figure}
    \centering
    \includegraphics[page=5,scale=0.75]{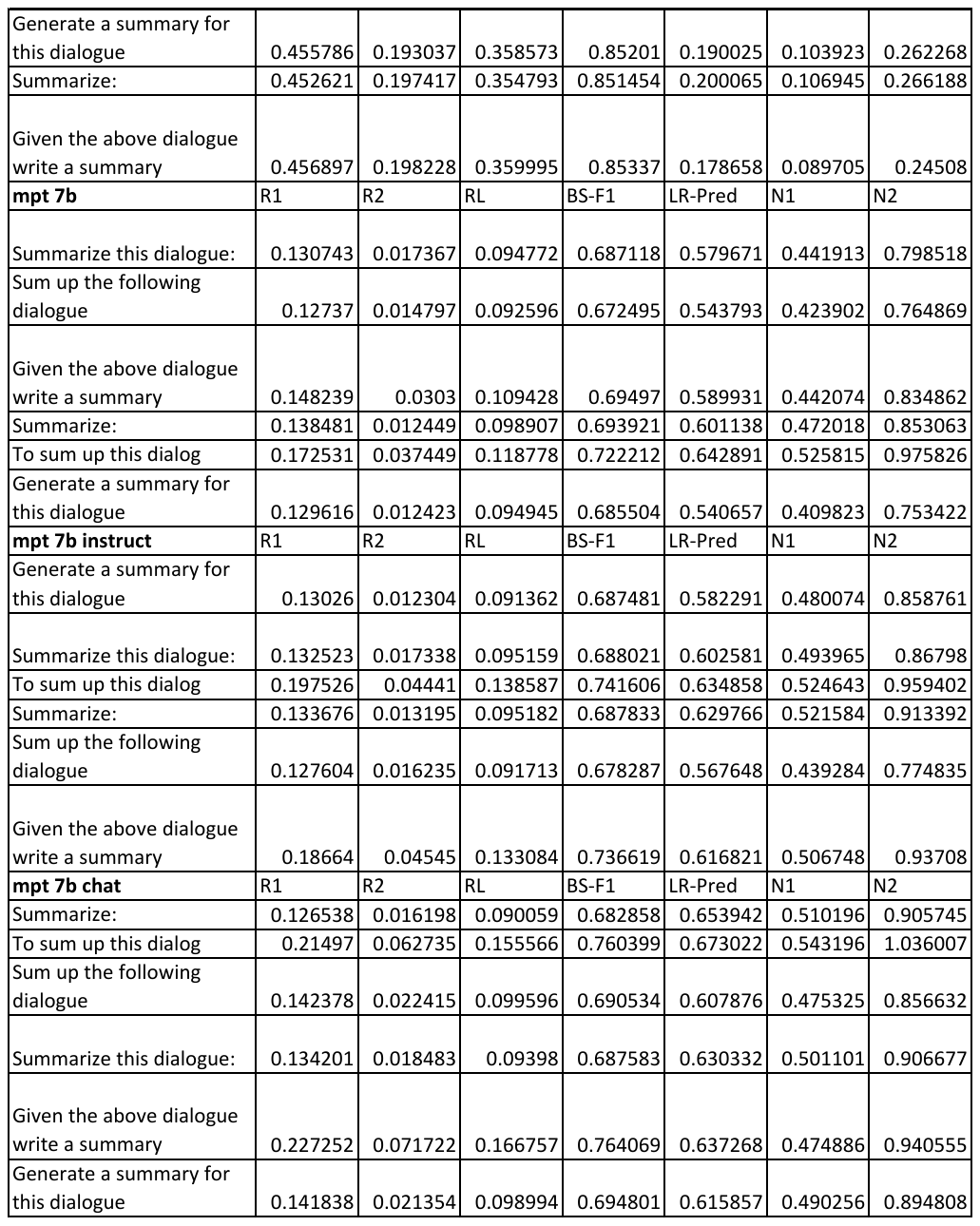}  
     \label{fig:samsum_prompts_5}
\end{figure}

\end{document}